\documentclass{article} 
\usepackage{iclr2022_conference,times}

\usepackage{amsmath,amsfonts,bm}

\def\Figref#1{Figure~\ref{#1}}

\def\eqref#1{equation~\ref{#1}}

\def\1{\bm{1}}

\def\vb{{\bm{b}}}

\def\vr{{\bm{r}}}

\def\vu{{\bm{u}}}
\def\vv{{\bm{v}}}

\def\mA{{\bm{A}}}

\def\mI{{\bm{I}}}

\def\mK{{\bm{K}}}

\def\mP{{\bm{P}}}
\def\mQ{{\bm{Q}}}

\def\mV{{\bm{V}}}
\def\mW{{\bm{W}}}
\def\mX{{\bm{X}}}

\DeclareMathAlphabet{\mathsfit}{\encodingdefault}{\sfdefault}{m}{sl}
\SetMathAlphabet{\mathsfit}{bold}{\encodingdefault}{\sfdefault}{bx}{n}

\newcommand{\R}{\mathbb{R}}

\definecolor{citecolor}{HTML}{0071bc}  
\usepackage[colorlinks=true, allcolors=citecolor]{hyperref}
\usepackage{url}

\usepackage{subcaption}
\usepackage{bbm}
\usepackage{booktabs}
\usepackage{multicol}
\usepackage{multirow}
\usepackage{color}
\usepackage{caption}
\usepackage{hhline}
\usepackage{pifont}
\usepackage{threeparttable}
\usepackage{makecell}
\usepackage{wrapfig}
\usepackage{colortbl}  
\usepackage{amssymb}

\usepackage[pdftex]{graphicx}  

\def\eg{\emph{e.g. }}

\def\etal{\emph{et al}.}

\newcommand{\xmark}{\ding{55}}
\newcommand{\cmark}{\ding{51}}

\definecolor{redstar}{RGB}{240,47,29}
\definecolor{curly}{RGB}{187,101,183}
\definecolor{poscolor}{RGB}{83,161,81}
\definecolor{negcolor}{RGB}{103,81,165}
\newcommand{\posval}[1]{{\textbf{\footnotesize\selectfont \color{poscolor}~($+$#1)}}}
\newcommand{\negval}[1]{{\textbf{\footnotesize\selectfont \color{negcolor}~($-$#1)}}}
\newcommand{\rebuttal}[1]{\textcolor{black}{#1}}

\newcommand{\modelname}{CycleMLP}
\newcommand{\longop}{Cycle Fully-Connected Layer}
\newcommand{\shortop}{Cycle FC}

\newcommand{\tabincell}[2]{\begin{tabular}{@{}#1@{}}#2\end{tabular}}

\title{CycleMLP: A MLP-like Architecture for Dense Prediction}

\author{
Shoufa~Chen\textsuperscript{1}
\quad Enze~Xie\textsuperscript{1}
\quad Chongjian~Ge\textsuperscript{1}
\quad Runjian~Chen\textsuperscript{1}
\quad Ding~Liang\textsuperscript{2}
\quad Ping~Luo\textsuperscript{1, 3} \\

\textsuperscript{1}~{The University of Hong Kong}
\quad \textsuperscript{2}~{SenseTime Research} \\
\textsuperscript{3}~{Shanghai AI Laboratory, Shanghai, China} \\

\tt\small{\{shoufach, xieenze, rhettgee, rjchen\}@connect.hku.hk} \\ \tt\small{liangding@sensetime.com} \quad \tt\small{pluo@cs.hku.hk}
}

\iclrfinalcopy 
\begin{document}

\maketitle

\begin{abstract}

This paper presents a simple MLP-like architecture, \modelname{}, which is a versatile backbone for visual recognition and dense predictions. As compared to modern MLP architectures, \eg, MLP-Mixer~\citep{mlp-mixer}, ResMLP~\citep{resmlp}, and gMLP~\citep{gmlp}, whose architectures are correlated to image size and thus are infeasible in object detection and segmentation,
\modelname{} has two advantages compared to modern approaches. 
(1) It can cope with various image sizes. 
(2) It achieves linear computational complexity to image size by using local windows. In contrast, previous MLPs have 
$O(N^2)$
computations due to fully spatial connections. 
We build a family of models which surpass existing MLPs and even state-of-the-art Transformer-based models, \eg Swin Transformer~\citep{liu2021Swin}, while using fewer parameters and FLOPs. 
We expand the MLP-like models' applicability, making them a versatile backbone for dense prediction tasks. \modelname{} achieves competitive results on object detection, instance segmentation, and semantic segmentation. In particular, \modelname{}-Tiny outperforms Swin-Tiny by 1.3\% mIoU on ADE20K dataset with fewer FLOPs. Moreover, CycleMLP also shows excellent zero-shot robustness on ImageNet-C dataset.
Code is available at \url{https://github.com/ShoufaChen/CycleMLP}.
\end{abstract}

\section{Introduction}\label{sec:introduction}

Vision models in computer vision have been long dominated by convolutional neural networks~(CNNs)~\citep{krizhevsky2012imagenet, he2016deep}. Recently, inspired by the successes in Natural Language Processing~(NLP) field, Transformers~\citep{vaswani2017attention} are adopted into the computer vision community. Built with self-attention layers, multi-layer perceptrons~(MLPs), and skip connections, Transformers make numerous breakthroughs on visual tasks~\citep{vit, liu2021Swin}.
More recently,~\citep{mlp-mixer, gmlp} have validated that building models solely on MLPs and skip connections without the self-attention layers can achieve surprisingly promising results on ImageNet~\citep{deng2009imagenet} classification. 

\definecolor{lightorange}{RGB}{246,165,114}
\begin{figure}[ht]
\begin{center}
\begin{minipage}[t]{0.35\textwidth}
\vspace{0pt}
\begin{subfigure}{\textwidth}
    \centering
    \includegraphics[width=\textwidth]{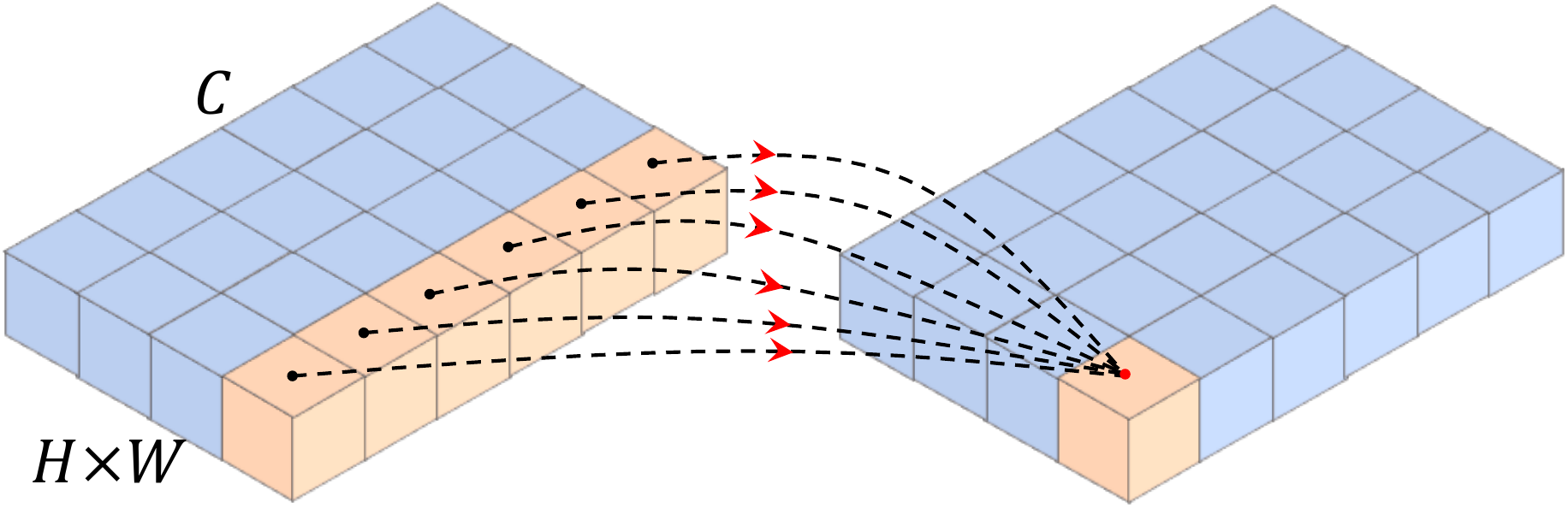}
    \vspace{-8mm}
    \caption{\scriptsize{Channel FC}}\label{fig:teaser-a}
    \vspace{-1mm}
\end{subfigure}
\begin{subfigure}{\textwidth}
    \centering
    \includegraphics[width=\textwidth]{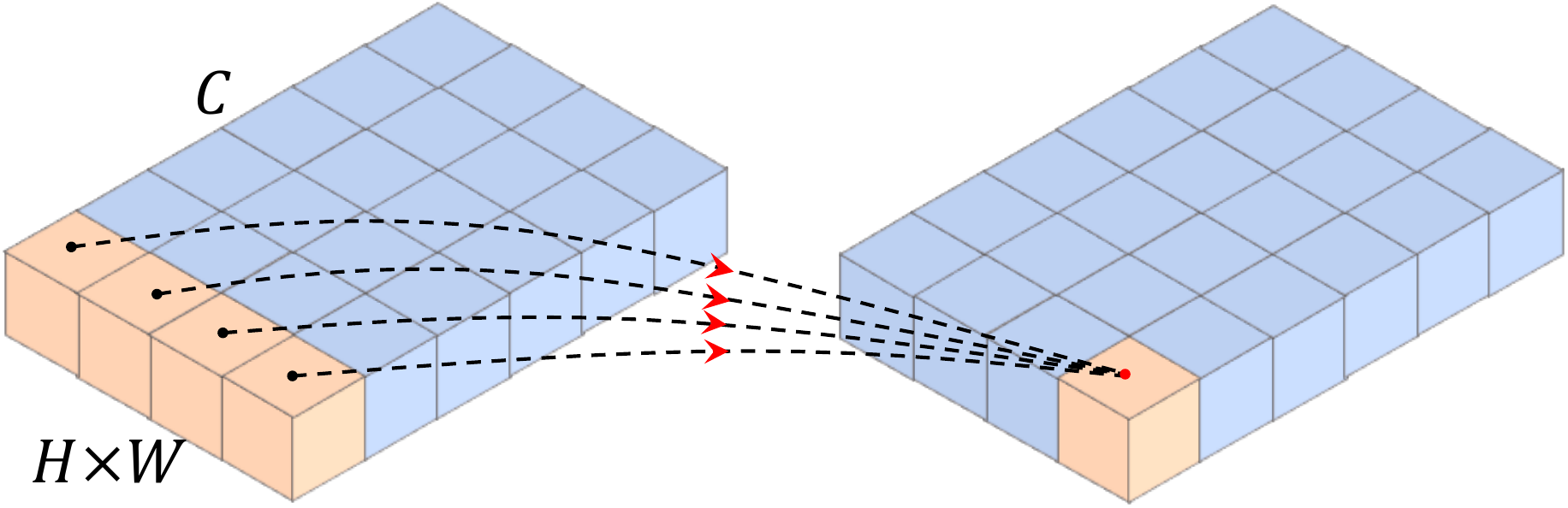}
    \vspace{-8mm}
    \caption{\scriptsize{Spatial FC}}\label{fig:teaser-b}
    \vspace{-1mm}
\end{subfigure}
\begin{subfigure}{\textwidth}
    \centering
    \includegraphics[width=\textwidth]{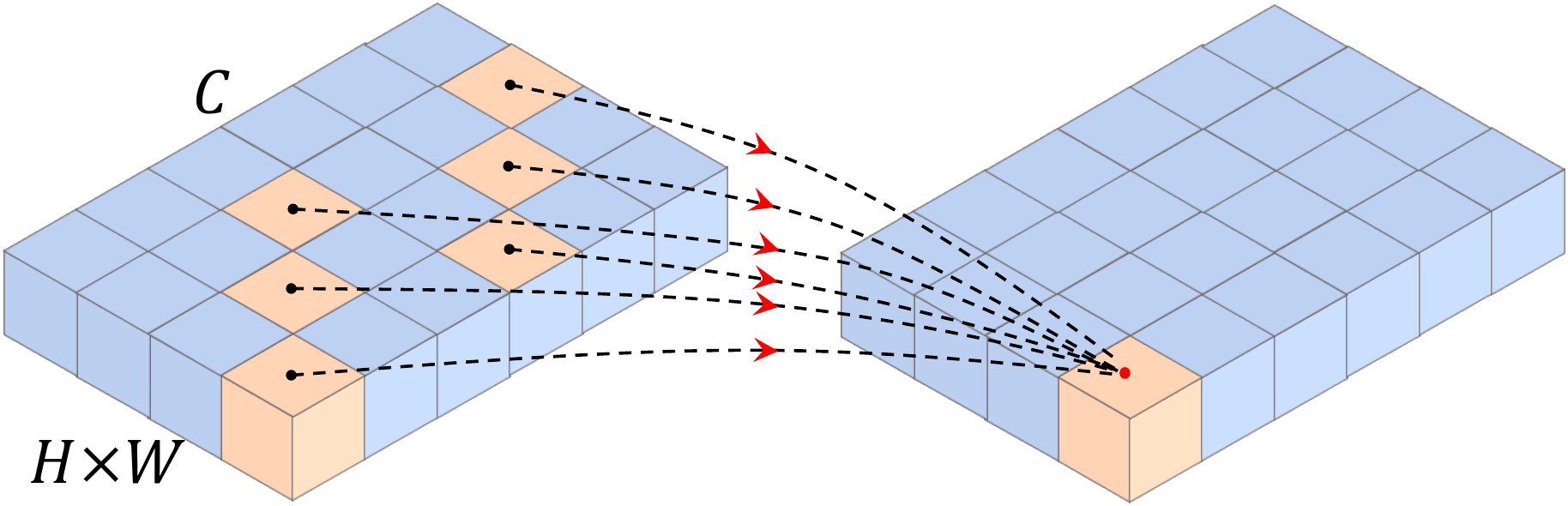}
    \vspace{-8mm}
    \caption{\scriptsize{Cycle FC}}\label{fig:teaser-c}
\end{subfigure}
\end{minipage}
\begin{minipage}[t]{0.63\textwidth}
\vspace{0pt}
\centering
\includegraphics[width=0.9\textwidth]{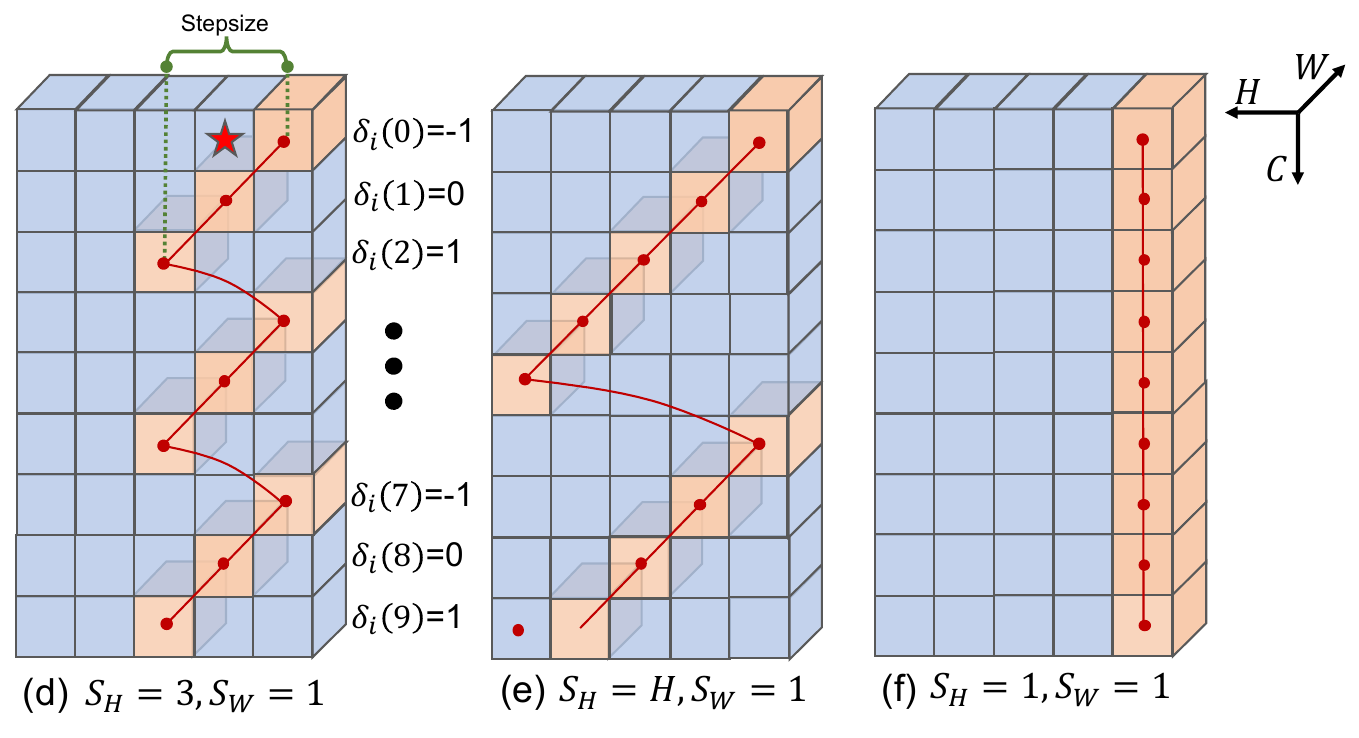}
\end{minipage}
\caption{\textbf{(a)-(c): motivation of \longop{} (Cycle FC)} compared to Channel FC and Spatial FC. \textbf{(a)}~Channel FC aggregates features in the channel dimension with spatial size `1'. It can handle various input scales but cannot learn spatial context. \textbf{(b)}~Spatial FC~\citep{mlp-mixer, resmlp, gmlp} has a global receptive field in the spatial dimension. However, its parameter size is fixed and it has quadratic computational complexity to image scale. \textbf{(c)}~Our proposed \longop{}~(\shortop{}) has linear complexity the same as channel FC and a larger receptive field than Channel FC. \textbf{(d)-(f): Three examples of different stepsizes.} \textcolor{lightorange}{Orange} blocks denote the sampled positions. \textcolor{redstar}{$\bigstar$} denotes the output position.
\rebuttal{For simplicity, we omit batch dimension and set the feature's width to 1 here for example. Several more general cases can be found in \Figref{fig:general}~(Appendix~\ref{sec:appendix-general}).} Best viewed in color.}
\label{fig:teaser}
\end{center}
\end{figure}

\begin{wraptable}{R}{0.6\textwidth}
\vspace{-1ex}
\centering

\small
\tabcolsep=0.07cm
\begin{tabular}{@{}ccccccc@{}}
\toprule
\multirow{2}{*}{FC}
&\multirow{2}{*}{\footnotesize{Stepsize}}
&\multirow{2}{*}{$\mathcal{O}(HW)$ }
&\multirow{2}{*}{\begin{tabular}[c]{@{}c@{}}\footnotesize{Scale} \\ \footnotesize{Variable} \end{tabular}}
&\multirow{2}{*}{\begin{tabular}[c]{@{}c@{}}ImgNet\\ Top-1\end{tabular}}
&\multirow{2}{*}{\begin{tabular}[c]{@{}c@{}}COCO\\ AP \end{tabular}}
&\multirow{2}{*}{\begin{tabular}[c]{@{}c@{}}ADE20K\\ mIoU\end{tabular}}
\\ 
  & & & & \\  \midrule
Channel   & 1  & $HW$     & \cmark & 79.4 & 35.0 & 36.3 \\
\midrule
Spatial   & -  & $H^2W^2$ & \xmark & 80.9 & \xmark & \xmark \\
\midrule
Cycle     & 7  & $HW$     & \cmark & 81.6 & 41.7 & 42.4 \\
\bottomrule
\end{tabular}

\caption{Comparison of three types of FC operators.}
\label{tab:3fcs}
\vspace{-2.5ex}
\end{wraptable}

Despite promising results on visual recognition tasks, these MLP-like models can not be used in dense prediction tasks (\textit{e.g.,} object detection and semantic segmentation) due to the three challenges:
~\textbf{(1)}~Current models are composed of blocks with non-hierarchical architectures, which make the model infeasible to provide pyramid and high-resolution feature representations.
~\textbf{(2)}~Current models cannot deal with flexible input scales due to the Spatial FC as shown in \Figref{fig:teaser-b}. 
The spatial FC is configured by an image-size related  weight\footnote{We omit \textit{bias} here for discussion convenience.}.
Thus, this structure typically requires the input image with a fixed scale during both the training and inference procedure.
It contradicts the requirements of dense prediction tasks, which usually adopt a multi-scale training strategy~\citep{carion2020end} and different input resolutions in training and inference stages~\citep{lin2014microsoft, cordts2016cityscapes}.
~\textbf{(3)}~The computational and memory costs of the current MLP models are quadratic to input image sizes for dense prediction tasks (e.g., COCO benchmark~\citep{lin2014microsoft}).

To address the first challenge, we construct a hierarchical architecture to generate pyramid features. For the second and third issues, we propose a novel variant of fully connected layer, named as \textit{\longop{}~(\shortop{})}, as illustrated in \Figref{fig:teaser-c}. The \shortop{} is capable of dealing with various image scales and has linear computational complexity to image size.

Our \shortop{} is inspired by Channel FC layer illustrated in \Figref{fig:teaser-a}, which is designed for channel information communication~\citep{lin2013network, szegedy2015going, he2016deep, howard2017mobilenets}. The main merit of Channel FC lies in that it can deal with flexible image sizes since it is configured by image-size agnostic weight of $C_{\textit{in}}$ and $C_{\textit{out}}$. However, the Channel FC is infeasible to aggregate spatial context information due to its limited receptive field.

Our \shortop{} is designed to enjoy Channel FC's merit of taking input with arbitrary resolution and linear computational complexity while enlarging its receptive field for context aggregation. Specifically, \shortop{} samples points in a cyclical style along the channel dimension~(\Figref{fig:teaser-c}). In this way, \shortop{} has the same complexity~(both the number of parameters and FLOPs) as channel FC while increasing the receptive field simultaneously. To this end, we adopt \shortop{} to replace the Spatial FC for spatial context aggregation~(\textit{i.e.,} token mixing) and build a family of MLP-like models for both recognition and dense prediction tasks.

The contributions of this paper are as follows:
~\textbf{(1)}~We propose a new MLP-like operator, \shortop{}, which is computational friendly to cope with flexible input resolutions.
~\textbf{(2)}~We take the first attempt to build a family of hierarchical MLP-like architectures (CycleMLP) based on \shortop{} operator for dense prediction tasks.
~\textbf{(3)}~Extensive experiments on various tasks (e.g., ImageNet classification, COCO object instance detection, and segmentation, and ADE20K semantic segmentation) demonstrate that CycleMLP outperforms existing MLP-like models and is comparable to and sometimes better than CNNs and Transformers on dense predictions.

\noindent
\textbf{Related Work.}
Convolution Neural Networks~(CNNs) has dominated the visual backbones for several years~\citep{krizhevsky2012imagenet, simonyan2014very, he2016deep}. \citep{vit} introduced the first pure Transformer-based~\citep{vaswani2017attention} model into computer vision and achieved promising performance, especially pre-trained on the large scale JFT dataset. Recently, some works~\citep{mlp-mixer, resmlp, gmlp} removed the attention in Transformer and proposed pure MLP-based models. 
Please see Appendix~\ref{sec:review} for a comprehensive review of the literature on the visual backbones.

\begin{figure}[t]
\centering
\begin{minipage}[t]{0.50\textwidth}
\vspace{0pt}
\includegraphics[width=\textwidth]{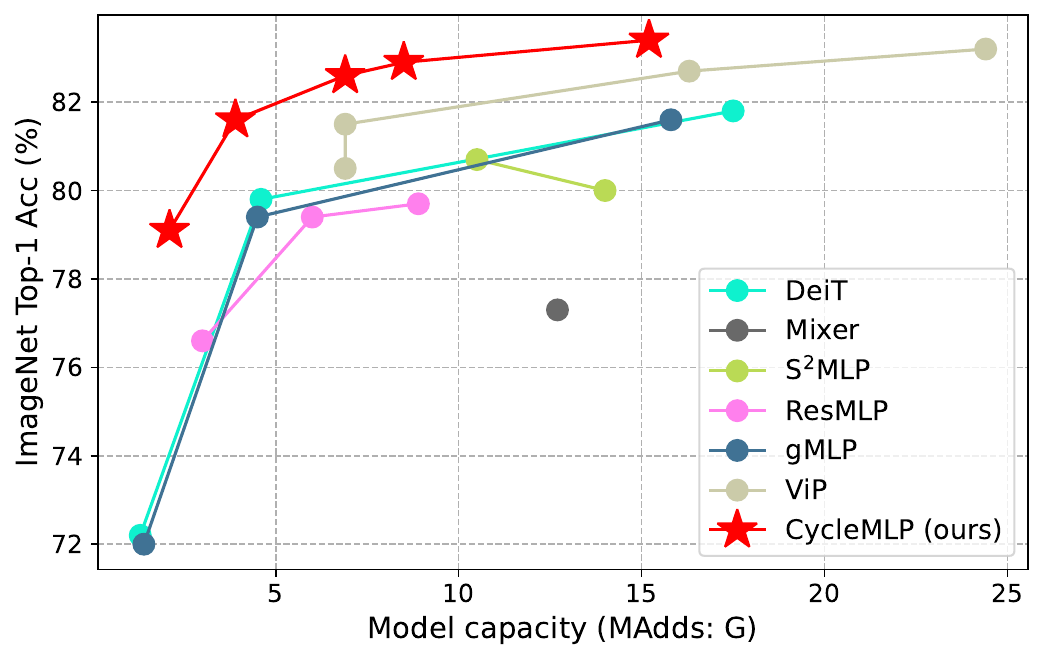}
\vspace{-5.5ex}
\end{minipage}
\hspace{0.03\textwidth}
\begin{minipage}[t]{0.4\textwidth}
\vspace{0pt}
\vspace{2pt}
\caption{\textbf{ImageNet accuracy \textit{v.s.} model capacity.} All models are trained on ImageNet-1K~\citep{deng2009imagenet} without extra data. \modelname{} surpasses existing MLP-like models such as MLP-Mixer~\citep{mlp-mixer}, ResMLP~\citep{resmlp}, gMLP~\citep{gmlp}, S$^2$-MLP~\citep{yu2021s} and ViP~\citep{hou2021vision}.}\label{fig:flops_acc}
\vspace{-5.5ex}
\end{minipage}
\end{figure}
\newcommand{\Mod}[1]{\ \mathrm{mod}\ #1}
\section{Method}

In this section, we introduce CycleMLP models for vision tasks including recognition and dense predictions. To begin with, in Sec.~\ref{subsection:operator} we formulate our proposed novel operator, \shortop{}, which serves as a basic component for building CycleMLP models. Then we compare \shortop{} with Channel FC and multi-head attention adopted in recent Transformer-based models~\citep{vit, touvron2020training, liu2021Swin} in Sec.~\ref{subsection:discussion}. Finally, we present the detailed configurations of CycleMLP models in Sec.~\ref{subsection:architecture}.

\subsection{\longop{}}\label{subsection:operator}
\textbf{Notation.} We denote an input feature map as $\mX \in \R^{H\times W\times C_{\textit{in}}}$, where $H, W$ denote the height and width of the image and $C_{\textit{in}}$ is the number of feature channels. We use subscripts to index the feature map. For example, $\mX_{i,j,c}$ is the value of $c^{th}$ channel at the spatial position $(i,j)$ and $\mX_{i,j,:}$ are values of all channels at the spatial $(i,j)$. 

\textbf{The motivation} behind \shortop{} is to enlarge receptive field of MLP-like models to cope with downstream dense prediction tasks while maintaining the computational efficiency. As illustrated in Figure~\ref{fig:teaser-a}, Channel FC applies weighting matrix on $\mX$ along the channel dimension on fixed position $(i,j)$. However, \shortop{} introduces a receptive field of $(S_H, S_W)$, 
where $S_H$ and $S_W$ are \textit{stepsize} along with the height and width dimension respectively~(illustrated in \Figref{fig:teaser}~(d)). The basic Cycle FC operator can be formulated as below:

\begin{equation}
\operatorname{\shortop{}}(\mX)_{i,j,:} = \sum_{c=0}^{C_{\textit{in}}}\mX_{i+\delta_i(c),j+\delta_j(c),c}\cdot\mW^{\text{mlp}}_{c,:}+\vb
\end{equation}

where $\mW^{\text{mlp}} \in \R^{C_{\textit{in}}\times C_{\textit{out}}}$ and $\vb \in \R^{C_{\textit{out}}}$ are parameters of \shortop{}. $\delta_i(c)$ and $\delta_j(c)$ are the spatial offset of the two axis on the $c^{th}$ channel, which are defined as below:

\begin{equation}
\delta_i(c)= (c \Mod{S_H}) -1,\quad\quad\quad \delta_j(c)= (\lfloor{\frac{c}{S_H}}\rfloor \Mod S_W) - 1
\end{equation}

\noindent\textbf{Examples.} We provide several examples~(Figure~\ref{fig:teaser}~(d)-(f)) to illustrate the stepsize. For the sake of visualization convenience, we set the tensor's $W=1$. Thus, these three examples naturally all have $S_W=1$.  Figure~\ref{fig:teaser}~(d) illustrates the offsets along two axis when $S_H=3$, that is $\delta_j(c) \equiv 0$ and $\delta_i(c)=\{-1,0,1,-1,0,1,\cdots\}$ when $c=0,1,2,\cdots,8$. Figure~\ref{fig:teaser}~(e) shows that when $S_H=H$, \shortop{} has a global receptive field. Figure~\ref{fig:teaser}~(f) shows that when $S_H=1$, there will be no offset along either axis and thus \shortop{} degrades to Channel FC~(\Figref{fig:teaser}~(a)). We also provide a more general case where $W\neq1$ and $S_H=3, S_W=3$ in Figure~\ref{fig:general}~(Appendix).

The offsets $\delta_i(c)$ and $\delta_j(c)$ enlarge the receptive field of \shortop{} as compared to Channel FC (\Figref{fig:teaser-a}), which applies weights solely on the same spatial position for all channels. The larger receptive field in return brings improvements on dense prediction tasks like semantic segmentation and object detection as shown in Table~\ref{tab:3fcs}. Meanwhile, \shortop{} still maintains computational efficiency and flexibility on input resolution. Both the FLOPs and the number of parameters are linear to the spatial scale which are exactly the same as those of Channel FC. In contrast, although Spatial FC has a global receptive field over the whole spatial space, its computational cost is quadratic to the image scale. Besides, it fails to handle inputs with different resolutions.

\subsection{Comparison between multi-head self-attention (MHSA) and \shortop{}}\label{subsection:discussion}

Inspired by \cite{Cordonnier2020On}, when re-parametried properly, a multi-head self-attention layer with $N_h$ heads can be formulated as below, which is similar to a convolution with kernel size $\sqrt{N_h}\times\sqrt{N_h}$. (Please refer to Appendix \ref{appendix: mhsa to conv} for detailed derivation)

\begin{equation}
\operatorname{MHSA}(\mX)_{i,j,:}=\sum_{h\in\{1,2,...,N_h\}}\mX_{i+\Delta_i(h),j+\Delta_j(h),:}\mW^{\text{mhsa},h} + \vb
\label{eq: mhsa in conv-like formulation}
\end{equation}

where $\mW^{\text{mhsa},h}\in\R^{C_{in}\times C_{out}}$ is the parameter matrix for $h^{th}$ head in $\text{MHSA}$. $\vb\in \R^{C_{out}}$ is the bias vector. $\{\Delta_i(h),\Delta_j(h)\}=\{(0,0),(1,0),(-1,0),\cdots\}$ contains all possible positional shift in convolution with kernel size $\sqrt{N_h}\times\sqrt{N_h}$. Further, we stack all $\mW^{\text{mhsa},h}$ together and reshape it into $\mW^{\text{mhsa}}\in\R^{K\times K\times C_{in}\times C_{out}}$. Then a relationship between $\mW^{\text{mlp}}$ and $\mW^{\text{mhsa}}$ can be formulated as follow. 

\begin{equation}
    \mW^{\text{mlp}}_{c,:} =\mW^{\text{mhsa}}_{\delta_i(c)+1,\delta_j(c)+1,c,:}
    \label{eq: mlp mhsa}
\end{equation}

\eqref{eq: mlp mhsa} shows that only the weights of $\mW^{\text{mhsa}}$ on spatial shift $(\delta_i(c)+1,\delta_j(c)+1)$ are taken into account in $\mW^{\text{mlp}}$. This indicates that \shortop{} introduce an inductive bias that the weighting matrix in $\operatorname{MHSA}$ should be sparse. Thus \shortop{} inherits the large receptive field introduced in $\operatorname{MHSA}$. The receptive field in \shortop{} is enlarged to $(S_H,S_W)$, which enables \shortop{} to tackle with downstream dense prediction tasks better. Meanwhile, with the sparsity inductive bias, \shortop{} maintains computational efficiency in MLP-based methods as compared to convolution and multi-head self-attention. The parameter size in \shortop{} is $C_{in}\times C_{out}$ while $\mW^{\text{mhsa}}\in\R^{K\times K\times C_{in}\times C_{out}}$.

\subsection{Overall Architecture}\label{subsection:architecture}

\noindent\textbf{Patch Embedding.}
Given the raw input image with the size of $H\times W\times 3$, our model first splits it into patches by a patch embedding module~\citep{vit}. Each patch is then treated as a ``token''. Specifically, we follow~\citep{fan2021multiscale, wang2021pvtv2} to adopt an overlapping patch embedding module with the window size 7 and stride 4. These raw patches are further projected to a higher dimension~(denoted as $C$) by a linear embedding layer. Therefore, the overall patch embedding module generates the features with the shape of $\frac{H}{4}\times\frac{W}{4}\times C$.

\noindent\textbf{CycleMLP Block.}
Then, we sequentially apply several \shortop{} Bloc blocks.
Comparing with the previous MLP blocks~\citep{mlp-mixer, resmlp, gmlp} visualized in \Figref{fig:detaild-mlp}~(Appendix), the key difference of \shortop{} block is that it utilizes our proposed \textit{\longop{}~(\shortop{})} for spatial projection and advances the models in context aggregation and information communication. Specifically, the \shortop{} block consists of three parallel \shortop{}s, which have stepsizes $S_H\times S_W$ of $1\times7$, $7\times1$, and $1\times1$. This design is inspired by the factorization of convolution~\citep{szegedy2016rethinking} and criss-cross attention~\citep{huang2019ccnet}. Then, there is a channel-MLP with two linear layers and a GELU~\citep{hendrycks2016gaussian} non-linearity in between. A LayerNorm~(LN)~\citep{ba2016layer} layer is applied before both parallel \shortop{} layers and channel-MLP modules. A residual connection~\citep{he2016deep} is applied after each module. 

\noindent\textbf{Stage.} The blocks with the same architecture are stacked to form one \textit{Stage}~\citep{he2016deep}. The number of tokens~(feature scale) is maintained within each stage. At each stage \textit{transition}, the channel capacity of the processed tokens is expanded while the number of tokens is reduced. This strategy effectively reduces the spatial resolution complexity. Overall, each of our model variants has four stages, and the output feature at the last stage has a shape of $\frac{H}{32}\times\frac{W}{32}\times C_4$. These stage settings are widely utilized in both CNN~\citep{simonyan2014very, he2016deep} and Transformer~\citep{wang2021pyramid, liu2021Swin} models. Therefore, CycleMLP can conveniently serve as a general-purpose visual backbone and a generic replacement for existing backbones.

\noindent\textbf{Model Variants.}
The design principle of the model's macro structure is mainly inspired by the philosophy of hierarchical Transformer~\citep{wang2021pyramid, liu2021Swin} models, which reduce the number of tokens at the transition layers as the network goes deeper and increase the channel dimension. In this way, we can build a hierarchical architecture that is critical for dense prediction tasks~\citep{lin2014microsoft, zhou2017scene}. 
Specifically, we build two model zoos following two widely used Transformer architectures, PVT~\citep{wang2021pyramid} and Swin~\citep{liu2021Swin}. Models in PVT-style are named from CycleMLP-B1 to CycleMLP-B5 and in Swin-Style are named as CycleMLP-T, -S, and -B, which represent models in \textit{tiny, small}, and \textit{base} sizes. 
These models are built by adapting several architecture-related hyper-parameters, including $S_i$, $C_i$, $E_i$, and $L_i$ which represent the stride of the transition, the token channel dimension, the number of blocks, and the expansion ratio respectively at Stage $i$. Detailed configurations of these models are in Table~\ref{tab:arch}~(Appendix). 

\section{Experiments}\label{sec:experiment}


\begin{table}[t]
    \centering
    \begin{minipage}[t]{0.42\textwidth}
    \vspace{0pt}
    \small
    \addtolength{\tabcolsep}{-2.pt}
    \newcommand{\arx}[2]{
\multirow{#1}{*}{arXiv:210#2}
}

\begingroup
\renewcommand{\arraystretch}{1.03}
\begin{tabular}{l | c c|c}
\Xhline{1.0pt}
Model & Param & FLOPs   &  Top-1 \\
\Xhline{1.0pt}
EAMLP-14         & 30M  & -    & 78.9 \\
EAMLP-19         & 55M  & -    & 79.4 \\ \hline

Mixer-B/16          & 59M  & 12.7G & 76.4  \\
Mixer-B/16$^\dagger$& 59M  & 12.7G & 77.3  \\ \hline

ResMLP-S12              & 15M  & 3.0G  & 76.6 \\
ResMLP-S24              & 30M  & 6.0G  & 79.4 \\
ResMLP-B24              & 116M & 23.0G & 81.0 \\ \hline

gMLP-Ti                   & 6M   & 1.4G  & 72.3 \\
gMLP-S                    & 20M  & 4.5G  & 79.6 \\
gMLP-B                    & 73M  & 15.8G & 81.6 \\ \hline

S$^2$-MLP-wide         & 71M  & 14.0G & 80.0 \\
S$^2$-MLP-deep         & 51M  & 10.5G & 80.7 \\ \hline

ViP-Small/7     & 25M   & 6.9G  & 81.5 \\
ViP-Medium/7     & 55M   & 16.3G & 82.7 \\
ViP-Large/7      & 88M   & 24.4G & \text{83.2} \\ \hline

AS-MLP-T           & 28M   & 4.4G  & 81.3 \\
AS-MLP-S           & 50M   & 8.5G  & 83.1 \\
AS-MLP-B           & 88M   & 15.2G & 83.3 
\\ \hline
\modelname{}-B1                        & 15M   & 2.1G  & 79.1 \\
\modelname{}-B2                        & 27M   & 3.9G  & 81.6 \\
\modelname{}-B3                        & 38M   & 6.9G  & 82.6 \\
\modelname{}-B4                        & 52M   & 10.1G & 83.0 \\
\modelname{}-B5                        & 76M   & 12.3G & 83.1 \\ 
\hline
\modelname{}-T                         & 28M   & 4.4G  & 81.3 \\
\modelname{}-S                         & 50M   & 8.5G  & 82.9 \\
\modelname{}-B                         & 88M   & 15.2G & \textbf{83.4} \\
\Xhline{1.0pt}
\end{tabular}
\endgroup
    \captionsetup{width=0.94\textwidth}
    \caption{ImageNet-1K classification for \textbf{MLP-like} models.}
    \label{tab:imagenet-mlp}
    \vspace{-4ex}
    \end{minipage}
    \hspace{0.02\textwidth}
    \begin{minipage}[t]{0.53\textwidth}
    \vspace{0pt}
    \small
    \addtolength{\tabcolsep}{-3.pt}
    \newcommand{\arx}[2]{
\multirow{#1}{*}{arXiv:210#2}
}

\newcommand{\STAB}[1]{\begin{tabular}{@{}c@{}}#1\end{tabular}}

\newcommand{\cnn}[1]{
\multirow{#1}{*}{\STAB{\rotatebox[origin=c]{90}{CNN}}}
}

\newcommand{\trans}[1]{
\multirow{#1}{*}{Transformer}
}

\begingroup
\renewcommand{\arraystretch}{1.05}
\begin{tabular}{l | c |c c c|c}
\Xhline{1.0pt}
Model & Family & Scale & Param & FLOPs   &  Top-1 \\
\Xhline{1.0pt}

ResNet18                   & CNN   & 224$^2$ & 12M & 1.8G  & 69.8 \\
EffNet-B3         & CNN   & 300$^2$ & 12M & 1.8G  & 81.6 \\
GFNet-H-Ti              & FFT   & 224$^2$ & 15M & 2.0G  & 80.1 \\
\modelname{}-B1                              & MLP   & 224$^2$ & 15M & 2.1G  & 78.9 \\ \hline

ResNet50                   &  CNN  & 224$^2$ & 26M & 4.1G  & 78.5 \\
DeiT-S           & Trans & 224$^2$ & 22M & 4.6G  & 79.8 \\
BoT-S1-50      & Hybrid& 224$^2$ & 21M & 4.3G  & 79.1 \\
PVT-S                & Trans & 224$^2$ & 25M & 3.8G  & 79.8 \\
Swin-T                   & Trans & 224$^2$ & 29M & 4.5G  & 81.3 \\
GFNet-H-S               & FFT   & 224$^2$ & 32M & 4.5G  & 81.5 \\
\modelname{}-B2         & MLP   & 224$^2$ & 27M & 3.9G  & 81.6 \\ \hline

ResNet101                 &  CNN  & 224$^2$ & 45M & 7.9G & 79.8 \\
RegNetY-8G  &  CNN  & 224$^2$ & 39M & 8.0G & 81.7 \\
BoT-S1-59     & Hybrid& 224$^2$ & 34M & 7.3G & 81.7  \\
PVT-M                 & Trans & 224$^2$ & 44M & 6.7G & 81.2 \\
\modelname{}-B3                              & MLP   & 224$^2$ & 38M & 6.9G & 82.4 \\ \hline

GFNet-H-B~        & FFT   & 224$^2$ & 54M & 8.4G  & 82.9 \\
Swin-S            & Trans & 224$^2$ & 50M & 8.7G  & 83.0 \\
PVT-L             & Trans & 224$^2$ & 61M & 9.8G  & 81.7 \\
\modelname{}-S    & MLP   & 224$^2$ & 50M & 8.5G  & 82.9 \\ \hline

ViT-B/16        & Trans & 384$^2$ & 86M & 55.4G & 77.9 \\
DeiT-B          & Trans & 224$^2$ & 86M & 17.5G & 81.8 \\
DeiT-B          & Trans & 384$^2$ & 86M & 55.4G & 83.1 \\
Swin-B          & Trans & 224$^2$ & 88M & 15.4G & 83.3 \\
\modelname{}-B  & MLP   & 224$^2$ & 88M & 15.2G & 83.4 \\


\Xhline{1.0pt}
\end{tabular}
\endgroup

    \caption{\textbf{Comparison with SOTA models on ImageNet-1K without extra data.}}
    \label{tab:imagenet-sota}
    \vspace{-4ex}
    \end{minipage}
\end{table}

In this section, we first examine \modelname{} by conducting experiments on ImageNet-1K~\citep{deng2009imagenet} image classification. Then, we present a bunch of baseline models achieved by \modelname{} in dense prediction tasks, \textit{i.e.,} COCO~\citep{lin2014microsoft} object detection, instance segmentation, and ADE20K~\citep{zhou2017scene} semantic segmentation.

\subsection{ImageNet-1K Classification}

The experimental settings for ImageNet classification are mostly from DeiT~\citep{touvron2020training}, Swin~\citep{liu2021Swin}.
The detailed experimental settings for ImageNet classification can be found in Appendix~\ref{sec:appendix-imagenet}.

\noindent
\textbf{Comparison with MLP-like Models.}
We first compare \modelname{} with existing MLP-like models and the results are summarized in Table~\ref{tab:imagenet-mlp} and Figure~\ref{fig:flops_acc}. The accuracy-FLOPs tradeoff of \modelname{} consistently outperforms existing MLP-like models~\citep{mlp-mixer, resmlp, gmlp, guo2021beyond, yu2021s, hou2021vision} under a wide range of FLOPs, which we attribute to the effectiveness of our \shortop{}. Specifically, compared with one of the pioneering MLP work, \textit{i.e.,} gMLP~\citep{gmlp}, \modelname{}-B2 achieves the same top-1 accuracy~(81.6\%) as gMLP-B while reducing more than 3$\times$ FLOPs~(3.9G for \modelname{}-B2 and 15.8G for gMLP-B). Furthermore, compared with existing SOTA MLP-like model, \textit{i.e.,} ViP~\citep{hou2021vision}, our model \modelname{}-B utilizes less FLOPs~(15.2G) than ViP-Large/7~(24.4G, the largest one of ViP family) while achiving higher top-1 accuracy.

It is noted that all previous MLP-like models listed in Table~\ref{tab:imagenet-mlp} do not conduct experiments on dense prediction tasks due to the incapability of dealing with variable input scales, which is discussed in Sec.~\ref{sec:introduction}. However, \modelname{} solved this issue by adopting \shortop{}. The experimental results on dense prediction tasks are presented in Sec.~\ref{sec:coco} and Sec.~\ref{sec:ade20k}.

\noindent
\textbf{Comparison with SOTA Models.}
Table~\ref{tab:imagenet-sota} further compares \modelname{} with previous state-of-the-art CNN, Transformer and Hybrid architectures. It is interesting to see that \modelname{} models achieve comparable performance to Swin Transformer~\citep{liu2021Swin}, which is the state-of-the-art Transformer-based model. Specifically, \modelname{}-B achieves slightly better top-1 accuracy~(83.4\%) than Swin-B~(83.3\%) with similar parameters and FLOPs. GFNet~\citep{rao2021global} utilizes the fast Fourier transform~(FFT)~\citep{cooley1965algorithm} to learn spatial information and achieves similar performance as \modelname{} on ImageNet-1K classification. However, the architecture of GFNet is correlated with the input resolution, and extra operation~(parameter interpolation) is required when input scale changes, which may hurt the performance of dense predictions. We will thoroughly compare \modelname{} with GFNet in Sec.~\ref{sec:ade20k} on ADE20K.

\begin{table}[t]
\centering
\begin{minipage}[t]{0.48\textwidth}
\vspace{0pt}
\renewcommand{\arraystretch}{1.12}
\addtolength{\tabcolsep}{-3.5pt}
    \small
    \begin{tabular}{c c c | c  c |c }
     \Xhline{1.0pt}
     $1\times7$  & $7\times1$    & $1\times1$  & Params  & FLOPs & Top-1 Acc \\
     \Xhline{1.0pt}

     \cmark &        & \cmark &\multirow{3}{*}{24.5M}& \multirow{3}{*}{3.6G}          &  80.5 \\
            & \cmark & \cmark &           &           &  80.4 \\
     \cmark & \cmark &        &           &           &  81.3 \\
     \hline
     \cmark \cmark   &   & \cmark  & \multirow{2}{*}{26.8M}  & \multirow{2}{*}{3.9G} & 80.6  \\
                     & \cmark \cmark &  \cmark  &     &      &   80.5  \\
     \hline
     \cmark & \cmark & \cmark & 26.8M   &  3.9G    &  81.6 \\
     \Xhline{1.0pt}
\end{tabular}
    \caption{\textbf{Ablation on three parallel branches.} We adopt \modelname{}-B2 variant for this ablation study. Double check marks~(\cmark \cmark) denote two same branches.}
    \label{tab:3branchs}
\end{minipage}
\hspace{0.02\textwidth}
\begin{minipage}[t]{0.44\textwidth}
\vspace{0pt}

\begin{tabular}{c | c | c}
\Xhline{1.0pt}
\multirow{2}{*}{Stepsize} &
\multirow{2}{*}{\begin{tabular}[c]{@{}c@{}}ImgNet\\ Top-1\end{tabular}}
&\multirow{2}{*}{\begin{tabular}[c]{@{}c@{}}ADE20K\\ mIoU\end{tabular}} \\
    &   & \\
\Xhline{1.0pt}
3  & 81.6 & 42.4 \\
5  & 81.6~\posval{0.0} & 43.2~\posval{0.8} \\
\cellcolor{red!5}{7}  & \cellcolor{red!5}{81.6~\posval{0.0}} & \cellcolor{red!5}\textbf{43.9}~\posval{1.5} \\
9  & 81.5~\negval{0.1} & 43.2~\posval{0.8} \\
\Xhline{1.0pt}
\end{tabular}
\caption{\textbf{\textit{Stepsize} ablation:} \modelname{} achieves the highest mIoU on ADE20K when stepsize is 7. However, the stepsize has negligible influence on the ImageNet classification. }
\label{tab:pattern}
\end{minipage}
\vspace{-10pt}
\end{table}

\subsection{Ablation Study}

In this subsection, we conduct extensive ablation studies to analyze each component of our design. Unless otherwise stated, We adopt \modelname{}-B2 instantiation in this subsection.

\noindent
\textbf{\longop{}.} To demonstrate the advantage of the \shortop{}, we compare \modelname{}-B2 with two other baseline models equipped with channel FC and Spatial FC as spatial context aggregation operators, respectively. The differences of these operators are visualized in Figure~\ref{fig:teaser}, and the comparison results are shown in Table~\ref{tab:3fcs}. \modelname{}-B2 outperforms the counterparts built on both Spatial and Channel FC for ImageNet classification, COCO object detection, instance segmentation, and ADE20K semantic segmentation. The results validate that \shortop{} is capable of serving as a general-purpose, plug-and-play operator for spatial information communication and context aggregation.

Table~\ref{tab:3branchs} further details the ablation study on the structure of \modelname{} block. It is observed that the top-1 accuracy drops significantly after removing one of the three parallel branches, especially when discarding the 1$\times$7 or 7$\times$1 branch. 
To eliminate the probability that the fewer parameters and FLOPs cause the performance drop, we further use two same branches~(denoted as ``\cmark \cmark'' in Table~\ref{tab:3branchs}) and one 1$\times$1 branch to align the parameters and FLOPs. The accuracy still drops relative to \modelname{}, which further demonstrates the necessity of these three unique branches.

\noindent
\textbf{Resolution adaptability.}
One remarkable advantage of \modelname{} is that it can take arbitrary-resolution images as input without any modification. On the contrary, GFNet~\citep{rao2021global} needs to interpolate the learnable parameters on the fly when the input scale is different from the one for training. 
We compare the resolution adaptability by directly evaluating models at a broad spectrum of resolutions using the weight pre-trained on 224$\times$224, without fine-tuning.
\begin{figure}
    \centering
\begin{minipage}[t]{0.6\textwidth}
\vspace{-2ex}
\centering
\vspace{0pt}
\begin{subfigure}{0.48\textwidth}
    \centering
    \includegraphics[width=\textwidth]{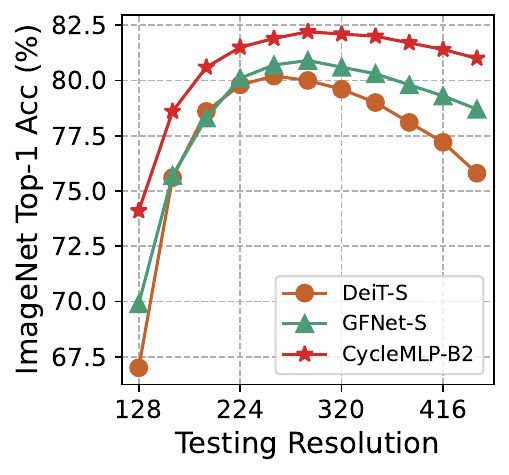}
\end{subfigure}
\hspace{-1mm}
\begin{subfigure}{0.48\textwidth}
    \centering
    \includegraphics[width=\textwidth]{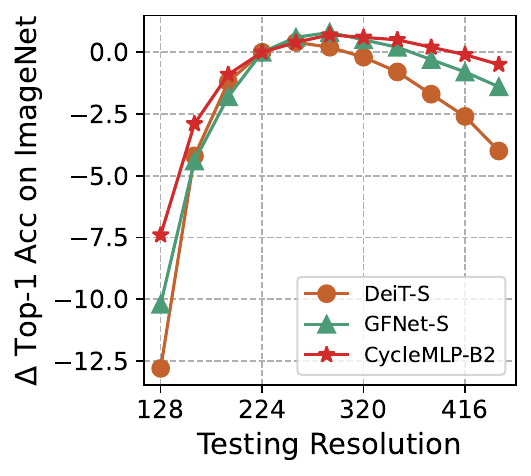}
\end{subfigure}
\end{minipage}
\begin{minipage}[t]{0.36\textwidth}
\vspace{0pt}
\caption{\small\textbf{Resolution adaptability.} All models are trained on 224$\times$224 and evaluated on various resolutions without fine-tuning. \textbf{Left:} Absolute top-1 accuracy; \textbf{Right:} Accuracy difference relative to that tested on 224$\times$224. The superiority of \modelname{}'s robustness becomes more significant when scale varies to a greater extent.}
\label{fig:adapt-resolution}
\end{minipage}
\end{figure}
Figure~\ref{fig:adapt-resolution}~(left) shows that the absolute Top-1 accuracy on ImagNet and Figure~\ref{fig:adapt-resolution}~(right) shows the accuracy differences between one specific resolution and the resolution of 224$\times$224. Compared with DeiT and GFNet, \modelname{} is more robust when resolution varies. In particular, at the 128$\times$128, \modelname{} saves more than 2 points drop compared to GFNet. Furthermore, at higher resolution, the performance drop of \modelname{} is less than GFNet. Note that the superiority of \modelname{} becomes more significant when the resolution changes to a greater extent.

\subsection{Object Detection and Instance Segmentation}\label{sec:coco}
\noindent\textbf{Settings.} We conduct object detection and instance segmentation experiments on COCO~\citep{lin2014microsoft} dataset. We first follow the experimental settings of PVT~\citep{wang2021pyramid}, which are introduced in Appendix.~\ref{sec:settings-pvt-coco}. The corresponding results are presented in Table~\ref{tab:coco}. Then, in order to compare fairly with Swin Transformer, which adopts a different experimental recipe with PVT, we further follow the experimental settings of Swin with our \modelname{}-S model and the results are presented in Table~\ref{tab:coco2}.

\begin{table}[h]
\centering
\renewcommand\arraystretch{1.05}
\setlength{\tabcolsep}{1.5pt}  
\footnotesize

\begin{tabular}{l|c |lcc|lcc| c|lcc|lcc}
\Xhline{1.0pt}
\renewcommand{\arraystretch}{0.05}
\multirow{2}{*}{Backbone} &\multicolumn{7}{c|}{RetinaNet 1$\times$} &\multicolumn{7}{c}{Mask R-CNN 1$\times$} \\
\cline{2-15} 
& Param & AP &AP$_{50}$ &AP$_{75}$ &AP$_S$ &AP$_M$ &AP$_L$ & Param & AP$^{\rm b}$ &AP$_{50}^{\rm b}$ &AP$_{75}^{\rm b}$  &AP$^{\rm m}$ &AP$_{50}^{\rm m}$ &AP$_{75}^{\rm m}$\\
\Xhline{1.0pt}
ResNet18        & 21.3M & 31.8 & 49.6 & 33.6 & 16.3 & 34.3 & 43.2 & 31.2M & 34.0 & 54.0 & 36.7 & 31.2 & 51.0 & 32.7 \\
PVT-Tiny        & 23.0M & 36.7 & 56.9 & 38.9 & 22.6 & 38.8 & 50.0 & 32.9M & 36.7 & 59.2 & 39.3 & 35.1 & 56.7 & 37.3 \\
\modelname{}-B1 & 24.9M & 38.1 & 58.7 & 40.1 & 21.9 & 41.9 & 50.4 & 34.8M & 39.8 & 61.7 & 43.3 & 37.0 & 58.8 & 39.7 \\
\hline

ResNet50        & 37.7M & 36.3 & 55.3 & 38.6 & 19.3 & 40.0 & 48.8 & 44.2M & 38.0 & 58.6 & 41.4 & 34.4 & 55.1 & 36.7 \\
PVT-Small       & 34.2M & 40.4 & 61.3 & 43.0 & 25.0 & 42.9 & 55.7 & 44.1M & 40.4 & 62.9 & 43.8 & 37.8 & 60.1 & 40.3 \\
\modelname{}-B2 & 36.5M & 40.6 & 61.4 & 43.2 & 22.9 & 44.4 & 54.5 & 46.5M & 42.1 & 64.0 & 45.7 & 38.9 & 61.2 & 41.8 \\
\hline

ResNet101        & 56.7M & 38.5 & 57.8 & 41.2 & 21.4 & 42.6 & 51.1 & 63.2M & 40.4 & 61.1 & 44.2 & 36.4 & 57.7 & 38.8 \\
ResNeXt101-32x4d & 56.4M & 39.9 & 59.6 & 42.7 & 22.3 & 44.2 & 52.5 & 62.8M & 41.9 & 62.5 & 45.9 & 37.5 & 59.4 & 40.2 \\
PVT-Medium       & 53.9M & 41.9 & 63.1 & 44.3 & 25.0 & 44.9 & 57.6 & 63.9M & 42.0 & 64.4 & 45.6 & 39.0 & 61.6 & 42.1 \\
\modelname{}-B3  & 48.1M & 42.5 & 63.2 & 45.3 & 25.2 & 45.5 & 56.2 & 58.0M & 43.4 & 65.0 & 47.7 & 39.5 & 62.0 & 42.4 \\ 
\hline

PVT-Large         & 71.1M & 42.6 & 63.7 & 45.4 & 25.8 & 46.0 & 58.4 & 81.0M & 42.9 & 65.0 & 46.6 & 39.5 & 61.9 & 42.5 \\
\modelname{}-B4                           & 61.5M & 43.2 & 63.9 & 46.2 & 26.6 & 46.5 & 57.4 & 71.5M & 44.1 & 65.7 & 48.1 & 40.2 & 62.7 & 43.5 \\
\hline

ResNeXt101-64x4d & 95.5M & 41.0 & 60.9 & 44.0 & 23.9 & 45.2 & 54.0 &101.9M & 42.8 & 63.8 & 47.3 & 38.4 & 60.6 & 41.3 \\
\modelname{}-B5                           & 85.9M & 42.7 & 63.3 & 45.3 & 24.1 & 46.3 & 57.4 & 95.3M & 44.1 & 65.5 & 48.4 & 40.1 & 62.8 & 43.0 \\
\Xhline{1.0pt}
\end{tabular}

\caption{\textbf{Object detection and instance segmentation on COCO \texttt{val2017}~\citep{lin2014microsoft}.} We compare \modelname{} with various backbones including ResNet~\citep{he2016deep}, ResNeXt~\citep{xie2017aggregated} and PVT~\citep{wang2021pyramid}.}
\label{tab:coco} 
\end{table}

\begin{table}[h]
\small
\centering
\scalebox{1.0}{\begin{tabular}{c | c c c | c c c | c c}
\toprule
Backbone & AP$^{b}$ & AP$^{b}_{50}$ & AP$^{b}_{75}$ & AP$^{m}$ & AP$^{m}_{50}$ & AP$^{m}_{75}$ & Params  & FLOPs  \tabularnewline
\midrule
ResNet50~\citep{he2016deep} & 41.0 & 61.7 & 44.9 & 37.1 & 58.4 & 40.1 & 44M & 260G  \tabularnewline
PVT-Small~\citep{wang2021pyramid} & 43.0 & 65.3 & 46.9 & 39.9 & 62.5 & 42.8 & 44M & 245G  \tabularnewline
Swin-T~\citep{liu2021Swin} & 46.0 & 68.2 & 50.2 & 41.6 & \textbf{65.1} & 44.8 & 48M & 264G  \tabularnewline
CycleMLP-T~\textbf{(ours)}   & \textbf{46.4} & \textbf{68.1} & \textbf{51.1} & \textbf{41.8} & 64.9 & \textbf{45.1} & 48M & 260G \\

\bottomrule

\end{tabular}}
\caption{The instance segmentation results of different backbones on the COCO val2017 dataset. Mask R-CNN frameworks are employed.}
\label{tab:coco2}
\end{table}

\noindent\textbf{Results.}
Firstly, as shown in Table~\ref{tab:coco}, \modelname{}-based RetinaNet consistently surpasses the CNN-based ResNet~\citep{he2016deep}, ResNeXt~\citep{xie2017aggregated} and Transformer-based PVT~\citep{wang2021pyramid} under similar parameter constraints, indicating that \modelname{} can serve as an excellent general-purpose backbone. Furthermore, using Mask R-CNN~\citep{he2017mask} for instance segmentation also demonstrates similar comparison results. Furthermore, from Table~\ref{tab:coco2}, the \modelname{} can achieve a slightly better performance than Swin Transformer.

\subsection{Semantic Segmentation}\label{sec:ade20k}
\noindent\textbf{Settings.} We conduct semantic segmentation experiments on ADE20K~\citep{zhou2017scene} dataset and present the detailed settings in Appendix.~\ref{sec:settings-pvt-ade20k}. Table~\ref{tab:ade20k} and Table~\ref{table:upper-lower} show the experimental results using training recipes from PVT and Swin respectively.

\begin{figure}[ht]
\centering
\begin{minipage}[t]{0.6\textwidth}
\vspace{0pt}
\addtolength{\tabcolsep}{-1pt}
\small
\begin{tabular}{l|c|c}
\Xhline{1.0pt}
	\multirow{2}{*}{Backbone} & \multicolumn{2}{c}{Semantic FPN}\\
	                          \cline{2-3}
	                          & Param & mIoU (\%)   \\
\Xhline{1.0pt}
	ResNet18~\citep{he2016deep}                & 15.5M & 32.9 \\
	PVT-Tiny~\citep{wang2021pyramid}           & 17.0M & 35.7 \\
	\modelname{}-B1~(ours)                    & 18.9M & \textbf{40.8} \\
	\hline

    ResNet50~\citep{he2016deep}                & 28.5M & 36.7 \\
    PVT-Small~\citep{wang2021pyramid}          & 28.2M & 39.8 \\
    Swin-T$^\dagger$~\citep{liu2021Swin}       & 31.9M & 41.5 \\
	GFNet-Tiny~\citep{rao2021global}           & 26.6M & 41.0 \\
	\modelname{}-B2~(ours)                    & 30.6M & \textbf{43.4} \\
    \hline
    ResNet101~\citep{he2016deep}               & 47.5M & 38.8\\
    ResNeXt101-32x4d~\citep{xie2017aggregated} & 47.1M & 39.7 \\
    PVT-Medium~\citep{wang2021pyramid}         & 48.0M & 41.6 \\
    GFNet-Small~\citep{rao2021global}          & 47.5M & 42.5 \\
    \modelname{}-B3~(ours)                    & 42.1M & \textbf{44.3} \\
    \hline
    PVT-Large~\citep{wang2021pyramid}          & 65.1M & 42.1 \\
    Swin-S$^\dagger$~\citep{liu2021Swin}       & 53.2M & \textbf{45.2} \\
    \modelname{}-B4~(ours)                    & 55.6M & 45.1 \\
    \hline
    GFNet-Base~\citep{rao2021global}           & 74.7M & 44.8 \\
    ResNeXt101-64x4d~\citep{xie2017aggregated} & 86.4M & 40.2 \\
    \modelname{}-B5~(ours)                    & 79.4M & \textbf{45.5} \\
\Xhline{1.0pt}
\end{tabular}
\captionsetup{width=0.92\textwidth}
\captionof{table}{\textbf{Semantic segmentation on ADE20K~\citep{zhou2017scene} val.} All models are equipped with Semantic FPN~\citep{kirillov2019panoptic}. $^\dagger$~Results are from GFNet~\citep{rao2021global}.}\label{tab:ade20k}
\end{minipage}
\begin{minipage}[t]{0.37\textwidth}
    \vspace{0pt}
    \begin{subfigure}{0.6\textwidth}
        \centering
        \includegraphics[width=\textwidth]{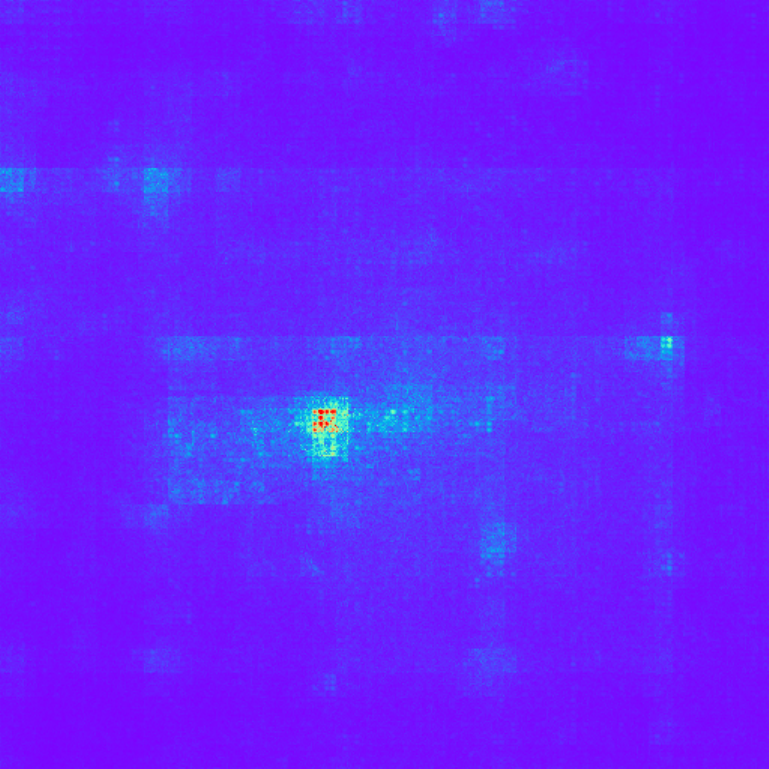}
        \vspace{-3ex}
        \caption{\footnotesize{Swin}}
    \end{subfigure}
    \begin{subfigure}{0.6\textwidth}
        \centering
        \includegraphics[width=\textwidth]{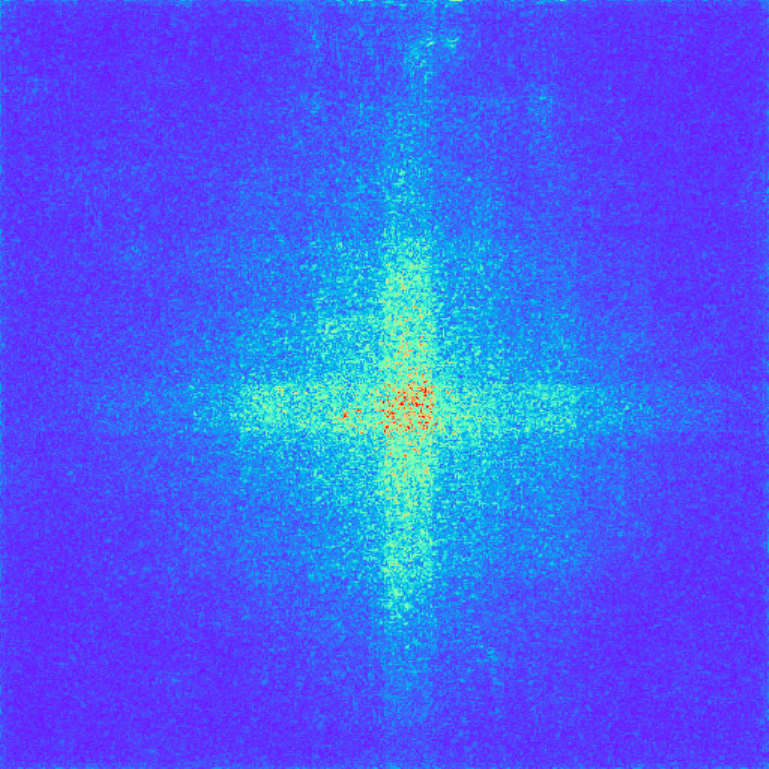}
        \vspace{-3ex}
        \caption{\footnotesize CycleMLP}
    \end{subfigure}
    \caption{\textbf{Effective Receptive Field (ERF)}. We visualize the ERFs of the last stage for both Swin~\citep{liu2021Swin} and CycleMLP. Best viewed with zoom in.}\label{fig:erf}
\end{minipage}
\begin{minipage}[t]{0.01\textwidth}
\vspace{0pt}
\centering
\hspace{-20\textwidth}
\includegraphics[height=50\textwidth]{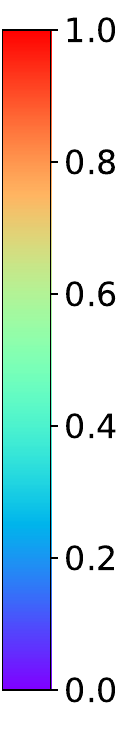}
\end{minipage}
\end{figure}

\begin{table}[h]
\begin{center}
\begin{tabular}{c| c| c| c| c}
\toprule
Method & Backbone & \tabincell{c}{val \\ MS mIoU}  & Params & FLOPs   \tabularnewline
\midrule
\multirow{3}{*}{UperNet~\citep{xiao2018unified}}
& Swin-T~\citep{liu2021Swin}   & 45.8 & 60M & 945G \tabularnewline
& AS-MLP-T~\citep{lian2021mlp} & 46.5 & 60M & 937G \tabularnewline  
& CycleMLP-T~(ours)            & \textbf{47.1} & 60M & 937G \tabularnewline
\midrule
\multirow{3}{*}{UperNet~\citep{xiao2018unified}}
& Swin-S~\citep{liu2021Swin}   & 49.5  & 81M & 1038G    \tabularnewline
& AS-MLP-S~\citep{lian2021mlp} & 49.2  & 81M & 1024G   \tabularnewline  
& CycleMLP-S~(ours)            & \textbf{49.6}  & 81M & 1024G \tabularnewline
\midrule
\multirow{3}{*}{UperNet~\citep{xiao2018unified}}
& Swin-B~\citep{liu2021Swin}  & 49.7 & 121M & 1188G    \tabularnewline 
& AS-MLP-B\citep{lian2021mlp} & 49.5  & 121M  & 1166G    \tabularnewline
& CycleMLP-B~(ours)           & \textbf{49.7}  & 121M  & 1166G   \tabularnewline
\bottomrule
\end{tabular}
\end{center}
\vspace{-12pt}
\caption{The semantic segmentation results of different backbones on the ADE20K validation set.}
\label{table:upper-lower}
\end{table}

\noindent
\textbf{Results.}
As shown in Table~\ref{tab:ade20k}, \modelname{} outperforms ResNet~\citep{he2016deep} and PVT~\citep{wang2021pyramid} significantly with similar parameters. Moreover, compared to the state-of-the-art Transformer-based backbone, Swin Transformer~\citep{liu2021Swin}, \modelname{} can obtain comparable or even better performance. Specifically, \modelname{}-B2 surpasses Swin-T by 0.9 mIoU with slightly less parameters~(30.6M \textit{v.s.} 31.9M).

Although GFNet~\citep{rao2021global} achieves similar performance as \modelname{} on ImageNet classification, \modelname{} notably outperforms GFNet on ADE20K semantic segmentation where input scale varies. We attribute the superiority of \modelname{} under a scale-variable scenario to the capability of dealing with arbitrary scales. On the contrary, GFNet~\citep{rao2021global} requires additional heuristic operation~(weight interpolation) when the input scale varies, which may hurt the performance. 

Moreover, we also visualized the receptive field following~\citep{xie2021segformer}, and the results are visualized in~\Figref{fig:erf}, which demonstrate that our \modelname{} has a  larger effective receptive field than Swin.

\subsection{Robustness}
\begin{table}[h]
    \centering
    \small
    \setlength\tabcolsep{1.5pt}
\begin{tabular}{@{}l c | c c c | c c c c | c c c  c | c c c c @{}}
\Xhline{2\arrayrulewidth}
\multirow{2}{*}{Network} & \multirow{2}{*}{mCE$\downarrow$} & \multicolumn{3}{c|}{Noise} & \multicolumn{4}{c|}{Blur} & \multicolumn{4}{c|}{Weather} & \multicolumn{4}{c}{Digital} \\
   &  & \scriptsize{Gauss} & \scriptsize{Shot} & \scriptsize{Impulse} & \scriptsize{Defocus} & \scriptsize{Glass} & \scriptsize{Motion} & \scriptsize{Zoom} & \scriptsize{Snow} & \scriptsize{Frost} & \scriptsize{Fog} & \scriptsize{Bright} & \scriptsize{Contrast} & \scriptsize{Elastic} & \scriptsize{Pixel} & \scriptsize{JPEG} \\
   
\Xhline{2\arrayrulewidth}
ResNet-50  & 76.7  & 79.8 & 81.6 & 82.6 & 74.7 & 88.6 & 78.0 & 79.9 & 77.8 & 74.8 & 66.1 & 56.6 & 71.4 & 84.7 & 76.9 & 76.8 \\
\hline
DeiT-S     & 54.6  & 46.3 & 47.7 & 46.4 & 61.6 & \textbf{71.9} & 57.9 & 71.9 & 49.9 & \textbf{46.2} & \textbf{46.0} & 44.9 & 42.3 & \textbf{66.6} & 59.1 & 60.4 \\
Swin-S     & 62.0  & 52.2 & 53.7 & 53.6 & 67.9 & 78.6 & 64.1 & 75.3 & 55.8 & 52.8 & 51.3 & 48.1 & 45.1 & 75.7 & 76.3 & 79.1 \\
\hline
MLP-Mixer  & 78.8  & 80.9 & 82.6 & 84.2 & 86.9 & 92.1 & 79.1 & 93.6 & 78.3 & 67.4 & 64.6 & 59.5 & 57.1 & 90.5 & 72.7 & 92.2 \\
ResMLP-12  & 66.0  & 57.6 & 58.2 & 57.8 & 72.6 & 83.2 & 67.9 & 76.5 & 61.4 & 57.8 & 63.8 & 53.9 & 52.1 & 78.3 & 72.9 & 75.3 \\
gMLP-S     & 64.0  & 52.1 & 53.2 & 52.5 & 73.1 & 77.6 & 64.6 & 79.9 & 77.7 & 78.8 & 54.3 & 55.3 & 43.6 & 70.6 & 58.6 & 67.5 \\
\hline
CycleMLP-S    & \textbf{53.7}  & \textbf{42.1} & \textbf{43.4} & \textbf{43.2} & \textbf{61.5} & 76.7 & \textbf{56.0} & \textbf{66.4} & \textbf{51.5} & 47.2 & 50.8 & \textbf{41.2} & \textbf{39.5} & 72.3 & \textbf{57.5} & \textbf{56.1} \\
\Xhline{2\arrayrulewidth}
\end{tabular}
    \caption{\textbf{Robustness on ImageNet-C~\citep{hendrycks2019robustness}.} The mean corruption error~(\textbf{mCE}) normalized by AlexNet~\citep{krizhevsky2012imagenet} errors is used as  the robustness metric. The lower, the better.}
    \label{tab:imagenet-c}
\end{table}

We further conduct experiments on ImageNet-C~\citep{hendrycks2016gaussian} to analyze the robustness ability of the CycleMLP, following~\citep{mao2021rethinking} and results are presented in Table~\ref{tab:imagenet-c}. Compared with both Transformers~(\eg DeiT and Swin) and existing MLP models~(\eg MLP-Mixer, ResMLP, gMLP), \modelname{} achieves a stronger robustness ability. 

\section{Conclusion}

We present a versatile MLP-like architecture, \modelname{}, in this work. \modelname{} is built upon the \longop{}~(\shortop{}), which is capable of dealing with variable input scales and can serve as a generic, plug-and-play replacement of vanilla FC layers. Experimental results demonstrate that \modelname{} outperforms existing MLP-like models on ImageNet classification and achieves promising performance on dense prediction tasks, \textit{i.e.,} object detection, instance segmentation and semantic segmentation. This work indicates that an attention-free architecture can also serve as a general vision backbone.

\noindent\textbf{Acknowledgment.} Ping Luo is supported by the General Research Fund of HK No.27208720, No.17212120, and the HKU-TCL Joint Research Center for Artificial Intelligence.

\bibliography{iclr2022_conference}
\bibliographystyle{iclr2022_conference}

\newpage
\appendix

\section{Literature on Vision Model}\label{sec:review}

\textbf{CNN-based Models.}
 Originally introduced over twenty years ago~\citep{lecun1989backpropagation}, convolutional neural networks~(CNN) have been widely adopted since the success of the AlexNet~\citep{krizhevsky2012imagenet} which outperformed prevailing approaches based on hand-crafted image features. There have been several attempts made to improve the design of CNN-based models. VGG~\citep{simonyan2014very} demonstrated a state-of-the-art performance on ImageNet via deploying small~($3\times 3$) convolution kernels to all layers. He \etal introduced skip-connections in ResNets~\citep{he2016deep}, enabling a model variant with more than 1000 layers. DenseNet~\citep{huang2017densely} connected each layer
to every other layer in a feed-forward fashion, strengthening feature propagation and reducing the number of parameters. In parallel with these architecture design works, some other works also made significant contributions to the popularity of CNNs, including normalization~\citep{ioffe2015batch, ba2016layer}, data augmentation~\citep{cubuk2020randaugment, yun2019cutmix, zhang2017mixup}, etc.

\noindent
\textbf{Transformer-based Models.}
Transformers were first proposed by Vaswani~\etal for machine translation and have since become the dominant choice in many NLP tasks~\citep{devlin2018bert, wang2018glue, yang2019xlnet, brown2020language}. Recently, transformer have also led to a series of breakthroughs in computer vision community since the invention of ViT~\citep{vit}, and have been working as a \textit{de facto} standard for various tasks, \textit{e.g.,} image classification~\citep{vit, touvron2020training, yuan2021tokens}, detection and segmentation~\citep{wang2021pyramid, liu2021Swin, zheng2021rethinking, xie2021segformer}, video recognition~\citep{wang2021end, bertasius2021space, arnab2021vivit, fan2021multiscale} and so on. Moreover, there has also been lots of interest in adopting transformer to cross aggregate multiple modality information~\citep{radford2021learning, gabeur2020multi, dzabraev2021mdmmt}. Furthermore, combining CNNs and transformers is also explored in~\citep{srinivas2021bottleneck, li2021bossnas, wu2021cvt, touvron2021going}.

\noindent
\textbf{MLP-based Models.} 
MLP-based models~\citep{mlp-mixer, resmlp, gmlp} differ from the above discussed CNN- and Transformer-based models because they resort to neither convolution nor self-attention layers. Instead, they use MLP layers over feature patches on spatial dimensions to aggregate the spatial context. 
These MLP-based models share similar macro structures but differ from each other in the detailed design of the micro block. In addition, MLP-based models provide more efficient computation than transformer-based models since they do not need to calculate affinity matrix using key-query multiplication. 
Concurrent to our work, S$^2$-MLP~\citep{yu2021s} utilizes a spatial-shift operation for spatial information communication. The similar aspect between our work and S$^2$-MLP lies in that we all conduct MLP operations along the channel dimension. However, our \shortop{} is different from S$^2$-MLP in: (1) S$^2$-MLP achieves communications between patches by splitting feature maps along channel dimension into several groups and shifting different groups in different directions. It introduces extra splitting and shifting operations on the feature map. On the contrary, we propose a novel operator-\longop{}-for spatial context aggregation. It does not modify the feature map and is formulated as a generic, plug-and-play MLP unit that can be used as a direct replacement of vanilla without any adjustments. (2) We design a pyramid structure for and conduct extensive experiments on classification, object detection, instance segmentation, and semantic segmentation. However, the output feature map of S$^2$-MLP has only one single scale in low resolution, which is unsuitable for dense prediction tasks. Only ImageNet classification is evaluated on S$^2$-MLP. We compared \shortop{} with S$^2$-MLP in details in the Section~\ref{sec:experiment}.

\rebuttal{}
\section{Comparison of MLP Blocks}

\begin{figure}[t]
\begin{center}
\begin{subfigure}{0.4\textwidth}
    \centering
    \includegraphics[width=\textwidth]{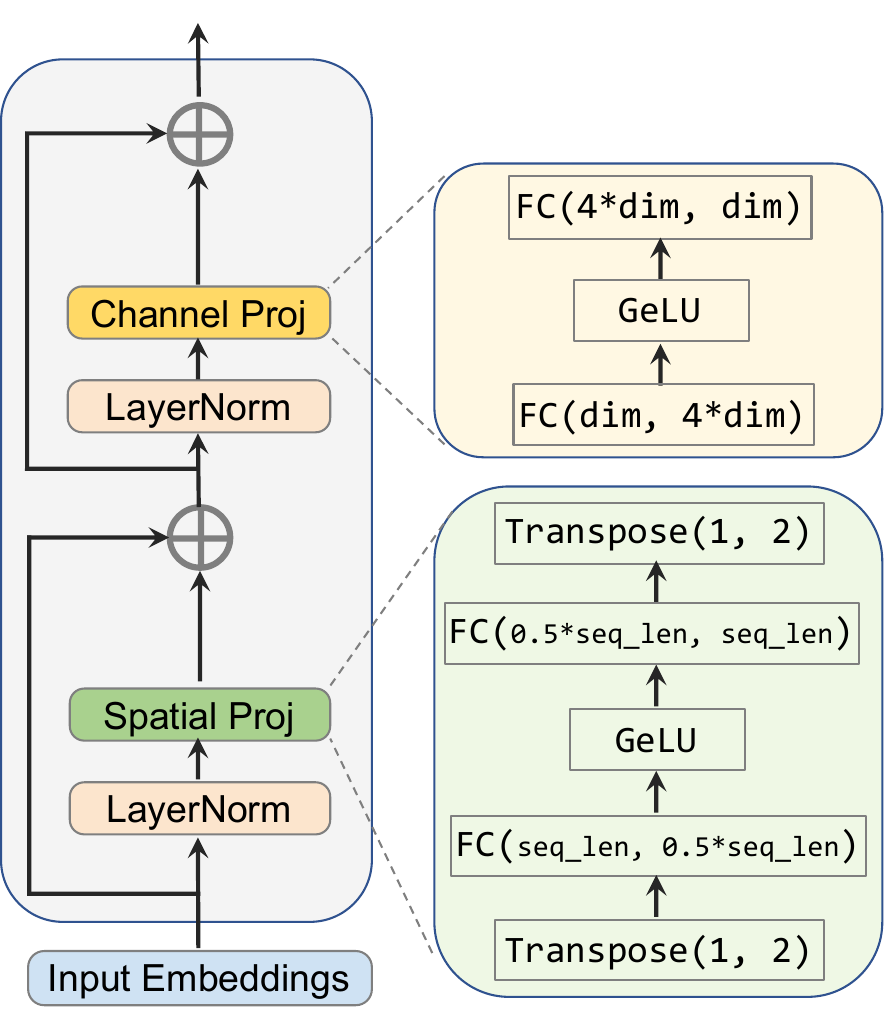}
    \caption{\text{MLP-Mixer~\citep{mlp-mixer}}}\label{fig:detail-mixer}
\end{subfigure}
\hspace{6mm}
\begin{subfigure}{0.4\textwidth}
    \centering
    \includegraphics[width=\textwidth]{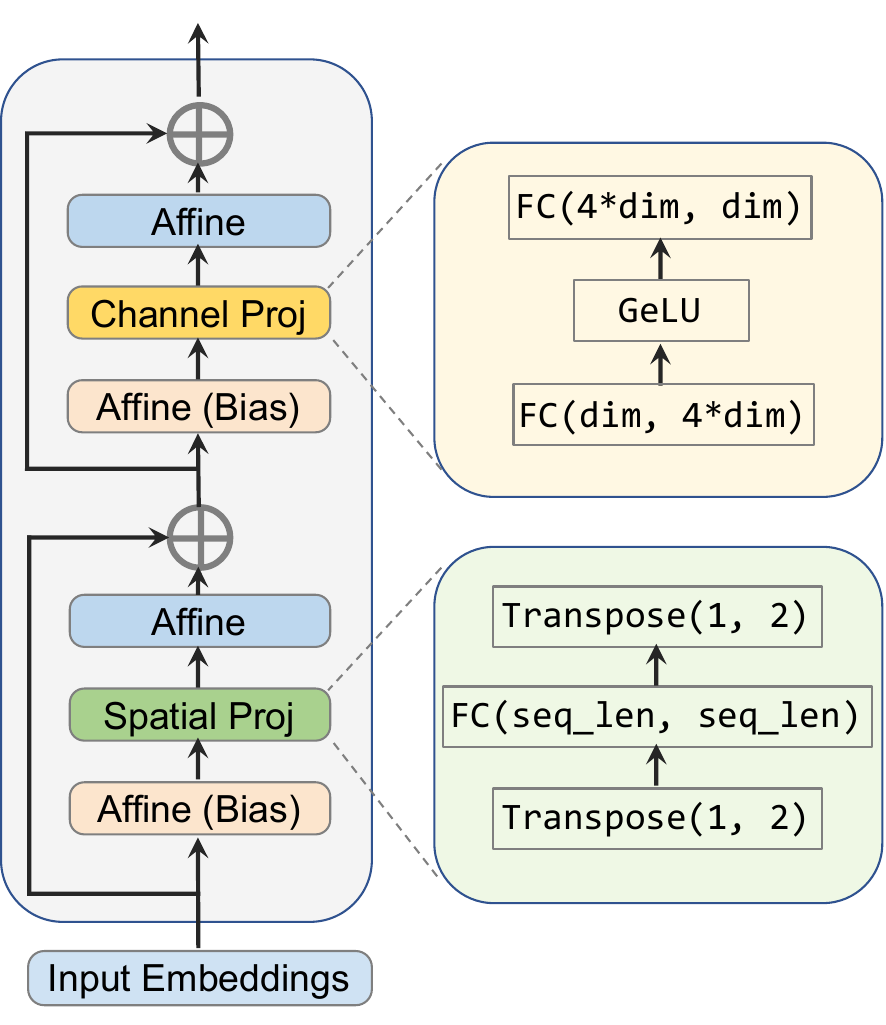}
    \caption{\text{ResMLP~\citep{resmlp}}}\label{fig:detail-resmlp}
\end{subfigure}
\begin{subfigure}{0.4\textwidth}
    \centering
    \includegraphics[width=\textwidth]{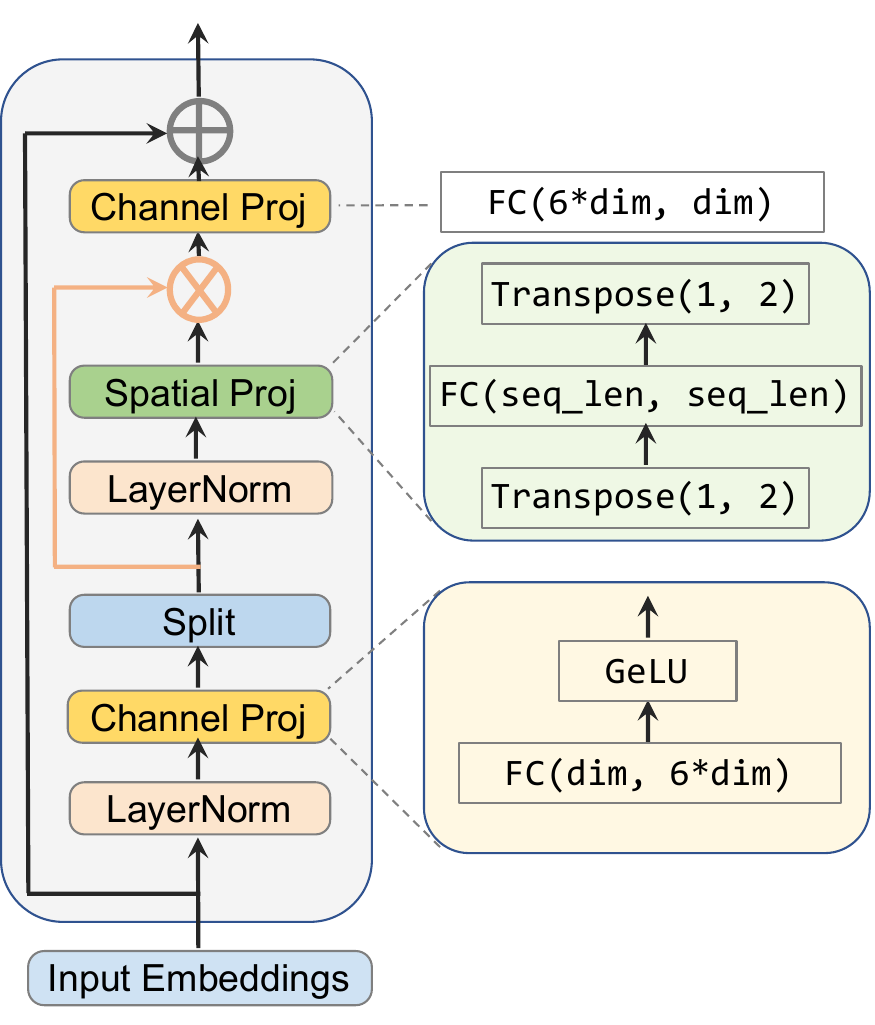}
    \caption{\text{gMLP~\citep{gmlp}}}\label{fig:detail-gmlp}
\end{subfigure}
\hspace{6mm}
\begin{subfigure}{0.4\textwidth}
    \centering
    \includegraphics[width=\textwidth]{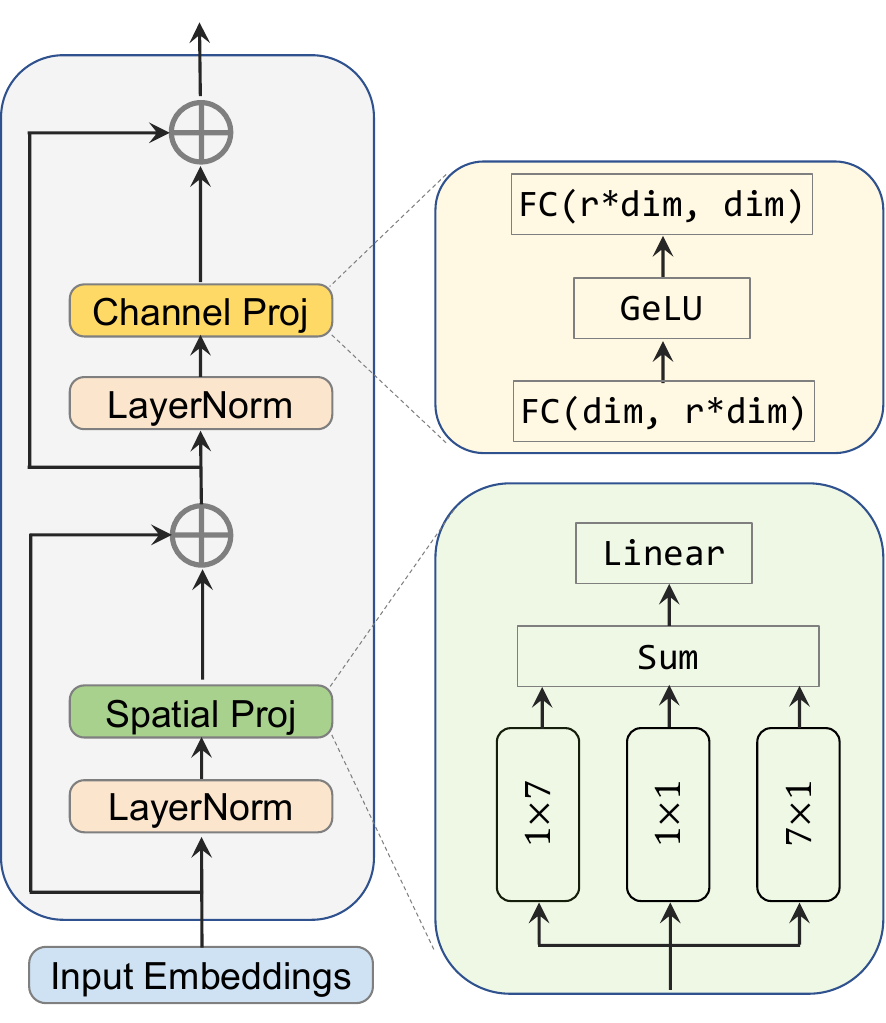}
    \caption{\text{CycleMLP~(Ours)}}\label{fig:detail-cyclemlp}
    \vspace{-1mm}
\end{subfigure}
\caption{Comparison of MLP blocks in details.}
\label{fig:detaild-mlp}
\end{center}
\end{figure}

We summary MLP blocks proposed by recent MLP-related works in Figure~\ref{fig:detaild-mlp}. We notice that existing MLP blocks, \textit{i.e.,} MLP-Mixer, ResMLP and gMLP share similar method of \textbf{Spatial Proj}: Transpose $\rightarrow$ Fully-Connected over spatial dimension $\rightarrow$ Transpose back. These models can not cope with variable image scales as the FC layers in Spatial Proj are configured by the \texttt{seq\_len}.

The blocks used for building \modelname{} consist of our proposed novel \shortop{}, whose configuration has nothing to do with image scales and can naturally deal with dynamic image scales.

\section{From multi-head self-attention to convolution}
\label{appendix: mhsa to conv}
In this section, we provide details in how $\operatorname{MHSA}$ can be transferred into a convolution-like operator in \eqref{eq: mhsa in conv-like formulation}. To start with, the a $\operatorname{MHSA}$ layer can be formulated as below:

\begin{equation}
\operatorname{MHSA}(\mX) = \operatornamewithlimits{concat}_{h\in\{1,...,N_{h}\}}[\operatorname{SA}_{h}(\mX)]\mW^{out}+\vb
\label{eq: mhsa}
\end{equation}

where $\mW^{out}\in\R^{(N_h C_{out})\times C_{out}'}$ and $\vb\in\R^{C_{out}'}$ are parameters for the final linear projection. $\operatorname{SA}_{h}$ is the $h^{th}$ self-attention module. Then we reshape $\mX$ into $\mX\in\R^{HW\times C_{in}}$ and let $T=H\times W$, which indicates that there are $T$ tokens in $\mX$. $\operatorname{SA}_{h}$ can be defined as follow:

\begin{equation}
\begin{split}
\operatorname{SA}(\mX)_{t,:} &= \operatorname{softmax}({\mA}_{t,:})\mV+\vb \\
\mA &=(\mQ+\mP)(\mK+\mP)^\intercal 
\end{split}
\label{eq: sa}
\end{equation}

where $\mV=\mX\mW^{\text{val}}$, $\mQ=\mX\mW^{\text{qry}}$, $\mK=\mX\mW^{\text{key}}$ are respectively the value, query and key matrix with learnable matrices $\mW^v\in \R^{C_{in}\times C_{out}}$, $\mW^q\in \R^{C_{in}\times C_{k}}$, $\mW^k\in \R^{C_{in}\times C_{k}}$. $\mP\in\R ^{T\times C_{in}}$ is the positional embedding matrix containing positional information for every input token, which can be replaced by the output of any function $f_{P}$ that encodes the position of tokens. And $\mA\in\R^{T\times T}$ is the attention matrix where each element $\mA_{i,j}$ is the attention score between the $i^{th}$ and $j^{th}$ token in $\mX$. With absolute positional encoding, the second line in \eqref{eq: sa} can be expanded as~\citep{Cordonnier2020On}:
\begin{small}
\begin{equation}
\begin{split}
  \mA_{q,k} =& (\mX_{q,:}+\mP_{q,:})\mW^{\text{qry}}((\mX_{k,:}+\mP_{k,:})\mW^{\text{key}})^\intercal \\ =&\mX_{q,:}\mW^{\text{qry}}(\mX_{k,:}\mW^{\text{key}})^\intercal+ \mX_{q,:}\mW^{\text{qry}}(\mP_{k,:}\mW^{\text{key}})^\intercal+\mP_{q,:}\mW^{\text{qry}}(\mX_{k,:}\mW^{\text{key}})^\intercal +\mP_{q,:}\mW^{\text{qry}}(\mP_{k,:}\mW^{\text{key}})^\intercal  
\end{split}
\end{equation}
\end{small}
When we apply relative positional encoding scheme in \citep{dai2019transformer}, $\mA$ is re-parametried into:

\begin{equation}
\mA_{q,k} = \mX_{q,:}\mW^{\text{qry}}(\mX_{k,:}\mW^{\text{key}})^\intercal+ \mX_{q,:}\mW^{\text{qry}}(\vr_{\delta_{q,k}}\hat{\mW}^{\text{key}})^\intercal+\vu(\mX_{k,:}\mW^{\text{key}})^\intercal+\vv(\vr_{\delta_{q,k}}\hat{\mW}^{\text{key}})^\intercal
\label{eq: relative positional encoding}
\end{equation}

where $\vr_{\delta_{q,k}}$ is a positional encoding for relative distance $\delta_{q,k}=(\delta_1,\delta_2)$ between token $q$ and $k$ in $\mX$. $\hat{\mW}^{key}$ is introduced to only pertain to the positional encoding $\vr_{q,k}$. $\vu$ and $\vv$ are learnable parameter vectors that replace the original $\mP_{q,:}\mW^{\text{qry}}$ term, which implies that the attention bias remains the same regardless of the absolution positions of the query. 
If we set $\mW^{qry}=\mW^{key}=0$ and $\hat{\mW}^{key}= \mI$, the first three terms in \eqref{eq: relative positional encoding} vanish and $\mA_{q,k} = \vv\vr_{q,k}^\intercal$. We set $\{\Delta_i(h),\Delta_j(h)\}=\{(0,0),(1,0),(-1,0),\cdots\}$ contains all possible positional shift in convolution with kernel size $\sqrt{N_h}\times\sqrt{N_h}$. For each head $h$, let $\vr_{q,k} = (\|\delta_{q,k}\|^2,\delta_1,\delta_2)$ and $v^h=-\alpha^h(1,-2\Delta_i(h),-2\Delta_j(h))$, each softmax attention matrix becomes: 

\begin{equation}
\operatorname{softmax}(\mA^h)_{q,k}=\left\{
\begin{aligned}
1&\ \ if\ \delta_{q,k}=(\Delta_i(h),\Delta_j(h)) \\
0&\ \ otherwise
\end{aligned}
\right.
\end{equation} 

Substitute $\operatorname{softmax}(\mA^h)$ into \eqref{eq: mhsa} and we get

\begin{equation}
\operatorname{MHSA}(\mX)_{i,j,:}=\sum_{h\in\{1,2,...,N_h\}}\mX_{i+\Delta_i(h),j+\Delta_j(h),:}\mW^{\text{mhsa},h} + \vb
\end{equation}

\section{Architecture Variants}

In order to conduct fair and convenient comparison, we build two model zoos: the one is in PVT-Style~(named as \modelname{}-B1 to -B5) and the other in Swin-Style~(named as \modelname{}-T, -S and -B). These models are scaled up by adapting several architecture-related hyper-parameters, including $S_i$, $C_i$, $E_i$ and $L_i$ which represent the stride of the transition, the token channel dimension, the number of blocks and the expansion ratio respectively at Stage $i$. Detailed configurations of these models are in Table~\ref{tab:arch}.

\begin{table}[t]
\small
\centering

\newcommand{\blocka}[2]{\multirow{3}{*}{\(\left[\begin{array}{c}\text{3$\times$3, #1}\\[-.1em] \text{3$\times$3, #1} \end{array}\right]\)$\times$#2}
}
\newcommand{\blockb}[3]{\multirow{3}{*}{\(\left[\begin{array}{c}\text{1$\times$1, #2}\\[-.1em] \text{3$\times$3, #2}\\[-.1em] \text{1$\times$1, #1}\end{array}\right]\)$\times$#3}
}

\newcommand{\outsize}[1]{\multirow{2}{*}[-2.5ex]{\scalebox{1.3}{$\frac{H}{#1}\times \frac{W}{#1}$}}}
\newcommand{\blockc}[4]{
$\begin{bmatrix}
	\begin{array}{l}
	R_1=#1 \\
	N_1=#2 \\
	E_1=#3 \\
	\end{array}
\end{bmatrix} \times #4$
}


\newcommand{\sblock}[3]{
$\begin{matrix}
E_{#1}=#2 \\
L_{#1}=#3 \\
\end{matrix}$
}

\newcolumntype{?}{!{\vrule width 1pt}}
\newcommand{\stitle}[6]{
\multirow{3}{*}[-1ex]{Stage #1} & \multirow{3}{*}[-1ex]{\scalebox{1.3}{$\frac{H}{#2}\times \frac{W}{#2}$}} & \multirow{2}{*}{\tabincell{c}{Overlapping\\Patch Embedding}} & \multicolumn{5}{c?}{$S_{#1}={#3}$} & \multicolumn{3}{c}{$S_{#1}={#3}$} \\
\cline{4-11}
    &    &    & \multicolumn{4}{c|}{$C_{#1}=#4$} & $C_{#1}=#5$  & \multicolumn{2}{c|}{$C_{#1}=#5$} & $C_{#1}=#6$ \\
\cline{3-11}
& & \tabincell{c}{\modelname{}\\Block}
}

\begingroup
\renewcommand{\arraystretch}{1.3}
\setlength{\tabcolsep}{1.6pt}
\scriptsize
\begin{tabular}{c|c|c|c|c|c|c|c ? c| c| c}
\bottomrule
   & \multirow{2}{*}{Output Size} & \multirow{2}{*}{Layer Name} & \multicolumn{5}{c?}{\text{PVT-Style}~\citep{wang2021pyramid}} & \multicolumn{3}{c}{Swin-Style~\citep{liu2021Swin}} \\
\cline{4-11}
 & & & B1 & B2 & B3 & B4 & B5  & Tiny  & Small & Base\\
\hline
\stitle{1}{4}{4}{64}{96}{128}    & \sblock{1}{4}{2} & \sblock{1}{4}{2}  & \sblock{1}{8}{3} &  \sblock{1}{8}{3}  & \sblock{1}{4}{3} & \sblock{1}{4}{2} & \sblock{1}{4}{2} & \sblock{1}{4}{2} \\
\hline
\stitle{2}{8}{2}{128}{192}{256}  & \sblock{2}{4}{2} & \sblock{2}{4}{3}  & \sblock{2}{8}{4} &  \sblock{2}{8}{8}  & \sblock{2}{4}{4} & \sblock{1}{4}{2} & \sblock{1}{4}{2} & \sblock{1}{4}{2} \\
\hline
\stitle{3}{16}{2}{320}{384}{512} & \sblock{3}{4}{4} & \sblock{3}{4}{10} & \sblock{3}{4}{18}&  \sblock{3}{4}{27} & \sblock{3}{4}{24} & \sblock{1}{4}{6} & \sblock{1}{4}{18} & \sblock{1}{4}{18} \\
\hline
\stitle{4}{32}{2}{512}{768}{1024} & \sblock{4}{4}{2} & \sblock{4}{4}{3}  & \sblock{4}{4}{3} &  \sblock{4}{4}{3}  & \sblock{4}{4}{3} & \sblock{1}{4}{2} & \sblock{1}{4}{2} & \sblock{1}{4}{2} \\
\hline
\multicolumn{3}{c|}{Parameters~(M)}& 15.2 & 26.8              &  38.4            &  51.8              &  75.7  & 28.3 & 49.6 & 87.7  \\
\hline
\multicolumn{3}{c|}{FLOPs~(G)}     & 2.1  &  3.9              &  6.9             &  10.1              &  12.3   &  4.4 & 8.6  & 15.2  \\
\toprule
\end{tabular}
\normalsize
\endgroup

\caption{\textbf{Instantiations of the \modelname{} with varying complexity.} The $E_i$ and $L_i$ denote the \textit{expand ratio} and \textit{number of repeated layers}. Our design principle is inspired by the philosophy of ResNet~\citep{he2016deep}, where the channel dimension increases while the spatial resolution shrinks with the layer going deeper.}
\label{tab:arch}
\end{table}

\section{Experimental setups}

\subsection{ImageNet Classification}\label{sec:appendix-imagenet}
\noindent
\textbf{Settings.} We train our models on the ImageNet-1K dataset~\citep{deng2009imagenet}, which contains 1.2M training images and 50K validation images evenly spreading 1,000 categories. We follow the standard practice in the community by reporting the top-1 accuracy on the validation set. Our code is implemented based on \texttt{PyTorch}~\citep{NEURIPS2019_9015} framework and heavily relies on the \texttt{timm}~\citep{rw2019timm} repository. For apple-to-apple comparison, our training strategy is mostly adopted from DeiT~\citep{touvron2020training}, which includes RandAugment~\citep{cubuk2020randaugment}, Mixup~\citep{zhang2017mixup}, Cutmix~\citep{yun2019cutmix} random erasing~\citep{zhong2020random} and stochastic depth~\citep{huang2016deep}. The optimizer is AdamW~\citep{loshchilov2017decoupled} with the momentum of 0.9 and weight decay of 5$\times$10$^{-2}$ by default. The cosine learning rate schedule is adopted with the initial value of 1$\times$10$^{-3}$. All models are trained for 300 epochs on 8 Tesla V100 GPUs with a total batch size of 1024.

Further kernel optimization for \shortop{} may bring a faster speed but is beyond the scope of this work. 

\subsection{COCO Instance Segmentation}\label{sec:settings-pvt-coco}
We conduct object detection and instance segmentation experiments on COCO~\citep{lin2014microsoft} dataset, which contains 118K and 5K images for \texttt{train} and \texttt{validation} splits. We adopt the \texttt{mmdetection}~\citep{mmdetection} toolbox for all experiments in this subsection. To evaluate the our \modelname{} backbones, we adopt two widely used detectors, \textit{i.e.,} RetinaNet~\citep{lin2017focal} and Mask R-CNN~\citep{he2017mask}. All backbones are initialized with ImageNet pre-trained weights and other newly added layers are initialized via Xavier~\citep{glorot2010understanding}. We use the AdamW~\citep{loshchilov2017decoupled} optimizer with the initial learning rate of 1$\times$10$^{-4}$. All models are trained on 8 Tesla V100 GPUs with a total batch size of 16 for 12 epochs~(\textit{i.e.,} 1$\times$ training scheduler). The input images are resized to the shorted side of 800 pixels and the longer side does not exceed 1333 pixels during training. We do not use the multi-scale~\citep{carion2020end, zhu2020deformable, sun2021sparse} training strategy. In the testing stage, the shorter side of input images is resized to 800 pixels while no constraint on the longer side.

\begin{table}[h]
\begin{center}
\begin{tabular}{c| c| c| c| c}
\toprule
Method & Backbone & \tabincell{c}{val \\ MS mIoU}  & Params & FLOPs   \tabularnewline
\midrule
DANet~\citep{fu2019dual} & ResNet-101 & 45.2 & 69M & 1119G   \tabularnewline
DeepLabv3+~\citep{chen2018encoder} & ResNet-101 & 44.1 & 63M & 1021G   \tabularnewline
ACNet~\citep{fu2019adaptive} & ResNet-101 & 45.9 & - & -  \tabularnewline
DNL \citep{yin2020disentangled} & ResNet-101 & 46.0 & 69M & 1249G \tabularnewline
OCRNet~\citep{yuan2020object} & ResNet-101 & 45.3 & 56M & 923G   \tabularnewline
UperNet~\citep{xiao2018unified} & ResNet-101 & 44.9 & 86M & 1029G  \tabularnewline
\midrule
OCRNet~\citep{yuan2020object} & HRNet-w48 & 45.7 & 71M & 664G   \tabularnewline
DeepLabv3+~\citep{chen2018encoder} & ResNeSt-101 & 46.9 & 66M & 1051G  \tabularnewline
DeepLabv3+~\citep{chen2018encoder} & ResNeSt-200 & 48.4 & 88M & 1381G 
\tabularnewline
\midrule
\multirow{3}{*}{UperNet~\citep{xiao2018unified}}
& Swin-T~\citep{liu2021Swin}   & 45.8 & 60M & 945G \tabularnewline
& AS-MLP-T~\citep{lian2021mlp} & 46.5 & 60M & 937G \tabularnewline  
& CycleMLP-T~(ours)            & \textbf{47.1} & 60M & 937G \tabularnewline
\midrule
\multirow{3}{*}{UperNet~\citep{xiao2018unified}}
& Swin-S~\citep{liu2021Swin}   & 49.5  & 81M & 1038G    \tabularnewline
& AS-MLP-S~\citep{lian2021mlp} & 49.2  & 81M & 1024G   \tabularnewline  
& CycleMLP-S~(ours)            & \textbf{49.6}  & 81M & 1024G \tabularnewline
\midrule
\multirow{3}{*}{UperNet~\citep{xiao2018unified}}
& Swin-B~\citep{liu2021Swin}  & 49.7 & 121M & 1188G    \tabularnewline 
& AS-MLP-B\citep{lian2021mlp} & 49.5  & 121M  & 1166G    \tabularnewline
& CycleMLP-B~(ours)           & \textbf{49.7}  & 121M  & 1166G   \tabularnewline
\bottomrule
\end{tabular}
\end{center}
\caption{The semantic segmentation results of different backbones on the ADE20K validation set.}
\label{table:upper}
\end{table}

\subsection{ADE20K Semantic Segmentation}\label{sec:settings-pvt-ade20k}

We conduct semantic segmentation experiments on ADE20K~\citep{zhou2017scene} dataset, which covers a broad range of 150 semantic categories. ADE20K contains 20K training, 2K validation and 3K testing images. We adopt the \texttt{mmsegmenation}~\citep{mmseg2020} toolbox as our codebase in this subsection. The experimental settings 
mostly follow PVT~\citep{wang2021pyramid}, which trains models for 40K iterations on 8 Tesla V100 GPUs with 4 samples per GPU. The backbone is initialized with the pre-trained weights on ImageNet. All models are optimized by AdamW~\citep{loshchilov2017decoupled}. The initial learning rate is configured as $2\times10^{-4}$ with the polynomial decay parameter of 0.9. Input images are randomly resized and cropped to 512$\times$512 at the training phase. During testing, we scale the images to the shorted side of 512. We adopt the simple approach Semantic FPN~\citep{kirillov2019panoptic} as the semantic segmentation method following~\citep{wang2021pyramid} for fair comparison.

\section{\rebuttal{Sampling Strategies}}

\begin{figure}[h]
\centering
\begin{minipage}[t]{0.3\textwidth}
\vspace{20pt}
 \includegraphics[width=0.98\textwidth]{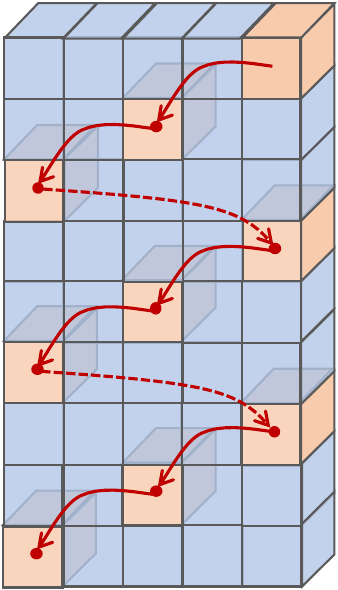}
    \caption{\rebuttal{\footnotesize An example of dilated CycleMLP where \texttt{dilation=2} and \texttt{stepsize=3}.}}\label{fig:pattern_dilation}
\end{minipage}
\hspace{0.08\textwidth}
\begin{minipage}[t]{0.48\textwidth}
\vspace{0pt}
\renewcommand{\arraystretch}{1.1}
\addtolength{\tabcolsep}{0.1pt}
\small
\begin{tabular}{c | c  c |c }
     \Xhline{1.0pt}
     Sampling  & Params  & FLOPs & Top-1 Acc \\
     \Xhline{1.0pt}
     \texttt{dilation=2} & 26.8M & 3.9G & 81.1 \\
     \hline
     Random, \texttt{S=1} &\multirow{3}{*}{26.8M}& \multirow{3}{*}{3.9G}  &  80.4 \\
     Random, \texttt{S=2} &           &           &  80.2 \\
     Random, \texttt{S=3} &           &           &  80.4 \\
     \hline
     \cellcolor{red!5}{CycleMLP} & \cellcolor{red!5}{26.8M}   &  \cellcolor{red!5}{3.9G}    &  \cellcolor{red!5}{\textbf{81.6}} \\
     \Xhline{1.0pt}
\end{tabular}
\vspace{-10pt}
\captionof{table}{\footnotesize \rebuttal{\textbf{Comparison with dilated and random sampling.} For random sampling, we conduct the experiments for three independent trials with three seeds~(\texttt{S=1, 2, 3}).}}\label{tab:random-sample}
\vspace{10pt}
\renewcommand{\arraystretch}{1.12}
\addtolength{\tabcolsep}{-3pt}
\small
\begin{tabular}{c | c | c  c |c }
     \Xhline{1.0pt}
     \scriptsize{Operators}  &  \scriptsize{Dense} & \scriptsize{Params}  & \scriptsize{FLOPs} & \scriptsize{Top-1 Acc} \\
     \Xhline{1.0pt}
     \scriptsize{Conv:} 1$\times$3 + 3$\times$1 & $\checkmark$ & 34.3M & 5.1G & 75.0 \\
     \scriptsize{CycleMLP:} 1$\times$3 + 3$\times$1  & \xmark & 26.8M & 3.9G & 76.1 \\
     \Xhline{1.0pt}
\end{tabular}
\vspace{-10pt}
\captionof{table}{\footnotesize \rebuttal{\textbf{Comparison with dense sampling:} On the consideration of training time, we only train both models for 100 epochs for fair comparison.}}\label{tab:dense-sampling}
\vspace{10pt}
\renewcommand{\arraystretch}{1.}
\addtolength{\tabcolsep}{5pt}
\small
\begin{tabular}{c  c | c | c}
\Xhline{1.0pt}
\multirow{2}{*}{branch1} & \multirow{2}{*}{branch2} &
\multirow{2}{*}{\begin{tabular}[c]{@{}c@{}}ImgNet\\ Top-1\end{tabular}}
&\multirow{2}{*}{\begin{tabular}[c]{@{}c@{}}ADE20K\\ mIoU\end{tabular}} \\
    &   & \\
\Xhline{1.0pt}
\cellcolor{red!5}{7$\times$1} & \cellcolor{red!5}{1$\times$ 7} & \cellcolor{red!5}{81.6} & \cellcolor{red!5}\textbf{43.9} \\
7$\times$2 & 2$\times$7 & 81.5 & 43.4 \\
7$\times$3 & 3$\times$7 & 81.4 & 42.7 \\
4$\times$4 & 4$\times$4 & 81.5 & 43.1 \\
\Xhline{1.0pt}
\end{tabular}
\vspace{-10pt}
\captionof{table}{\footnotesize \rebuttal{\textbf{Comparison on different stepsizes (e.g., even stepsize and odd stepsize)}, including 7$\times$2, 4$\times$4.}}\label{tab:even}
\end{minipage}
\end{figure}

\rebuttal{We explore more sampling strategies in this subsection, including random sampling and dilated sampling inspired by dilated convolution~\citep{YuKoltun2016, chen2018encoder}~(as shown in Figure~\ref{fig:pattern_dilation}). We also compare the dense sampling method with ours.}

\noindent\rebuttal{\textbf{Random sampling.} As shown in Table~\ref{tab:random-sample}, we conduct experiments with random sampling for three independent trials and observe that the averaged Top-1 accuracy on ImageNet-1K drops by $1.3\%$. We hypothesize that the decreased performance is caused by the fact that random sampling will totally disturb the semantic information of objects, which is essential to image recognition. Compared with the random sampling strategy, our cyclical sampling is able to aggregate the adjacent pixels, which benefits in capturing the semantic information.}

\noindent\rebuttal{\textbf{Dilated Stepsize}~(Figure~\ref{fig:pattern_dilation}). As shown in Table~\ref{tab:random-sample}, we observe the result of dilated sampling is better than the random one ($+1.0\%$ acc) but lower than ours ($-0.5\%$ acc). In fact, compared with the random sampling, dilated solutions take their advantages in local information aggregation. However, compared with the cyclical sampling strategy, dilated solutions lose the fine-grained information for recognition. It may hurt the accuracy performance to some extent.}

\noindent\rebuttal{\textbf{Dense sampling.} we conduct ablation studies by using dense sampling strategies (i.e., vanilla convolution with kernel size 1$\times$3 and 3$\times$1). Since dense sampling strategies incredibly increase the models’ parameters and FLOPs, we do not have enough time to thoroughly optimize the model for 300 epochs. Therefore, for fair comparisons, we conducted extra ablation studies on training models for 100 epochs with the strictly same learning configurations. The results shown in Table~\ref{tab:dense-sampling} demonstrate that the sparse sampling strategy (ours) outperforms the dense one. The comparison indicates that the dense sampling strategies introduce redundant parameters, which makes the model hard to optimize. Our sparse sampling strategy with fewer parameters is proven to be efficient and optimization-friendly.}

\section{\rebuttal{Visualization Examples}}\label{sec:appendix-general}

\begin{figure}
    \centering
\begin{subfigure}{0.46\textwidth}
    \centering
    \includegraphics[width=\textwidth]{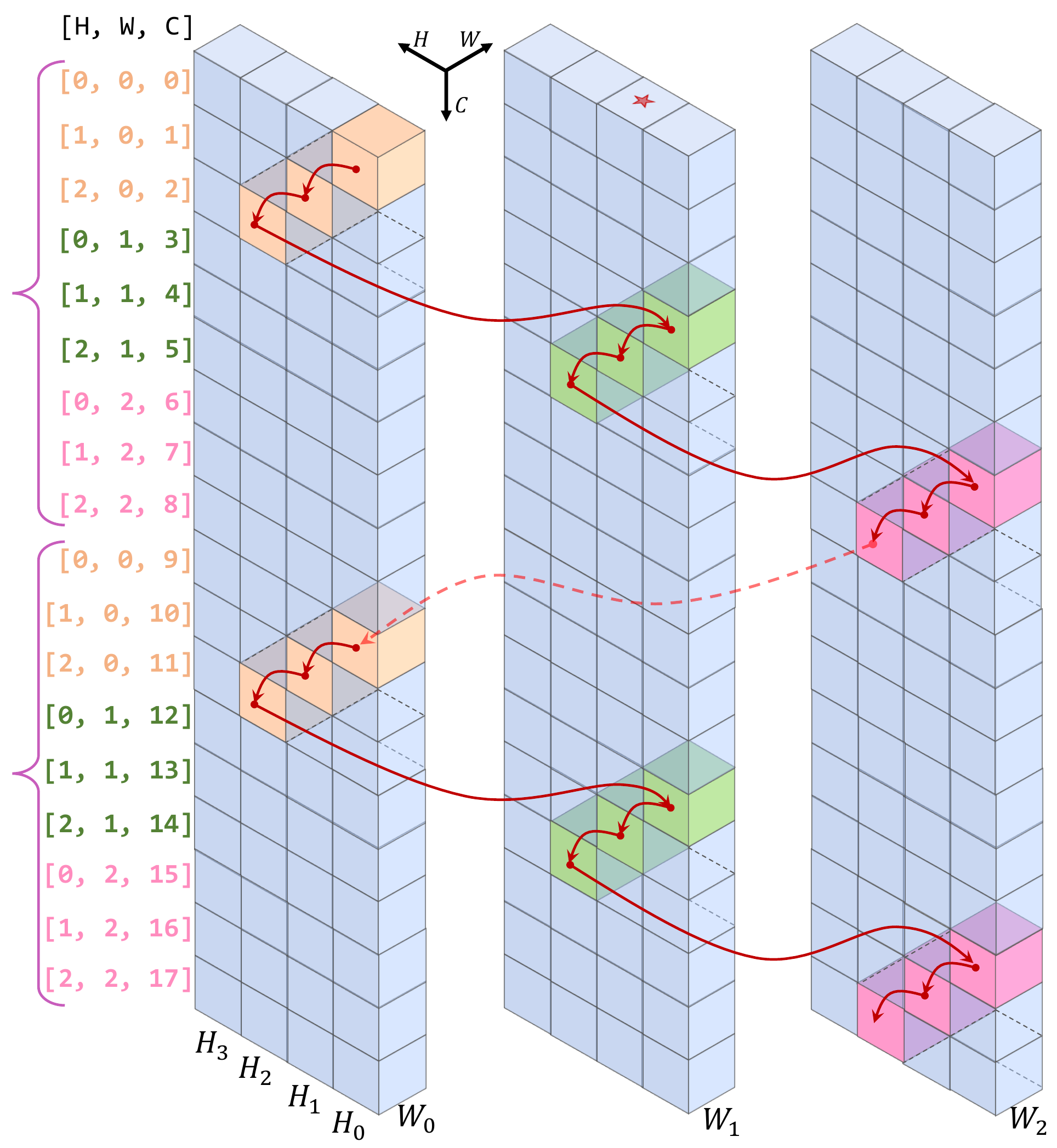}
    \vspace{-4mm}
    \caption{$S_H\times S_W$: $3\times3$}\label{fig:pattern3x3}
    \vspace{2mm}
\end{subfigure}
\hfill\vline\hfill
\begin{subfigure}{0.46\textwidth}
    \centering
    \includegraphics[width=\textwidth]{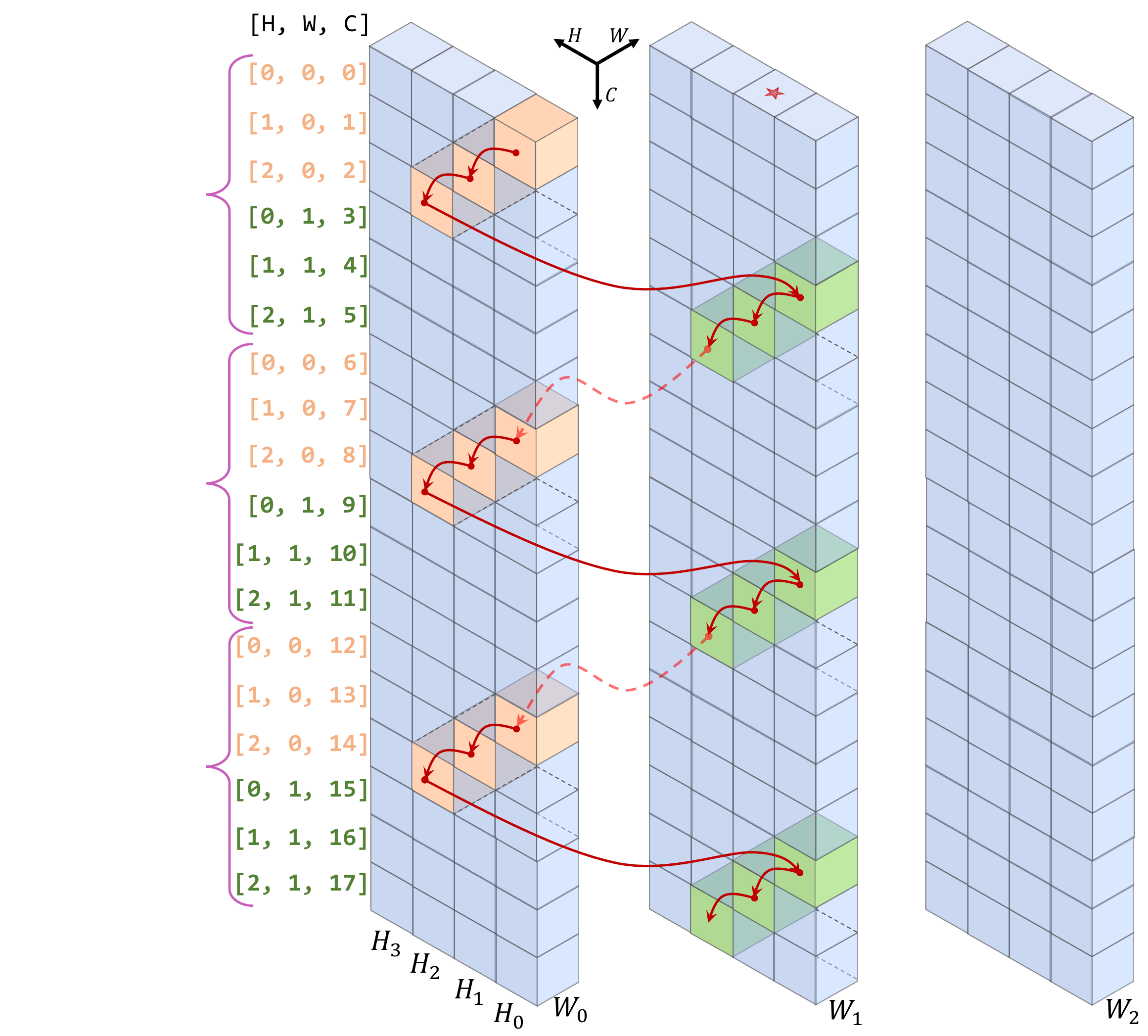}
    \vspace{-4mm}
    \caption{$S_H\times S_W$: $3\times2$}\label{fig:pattern3x2}
    \vspace{2mm}
\end{subfigure}

\begin{subfigure}{0.46\textwidth}
    \centering
    \includegraphics[width=\textwidth]{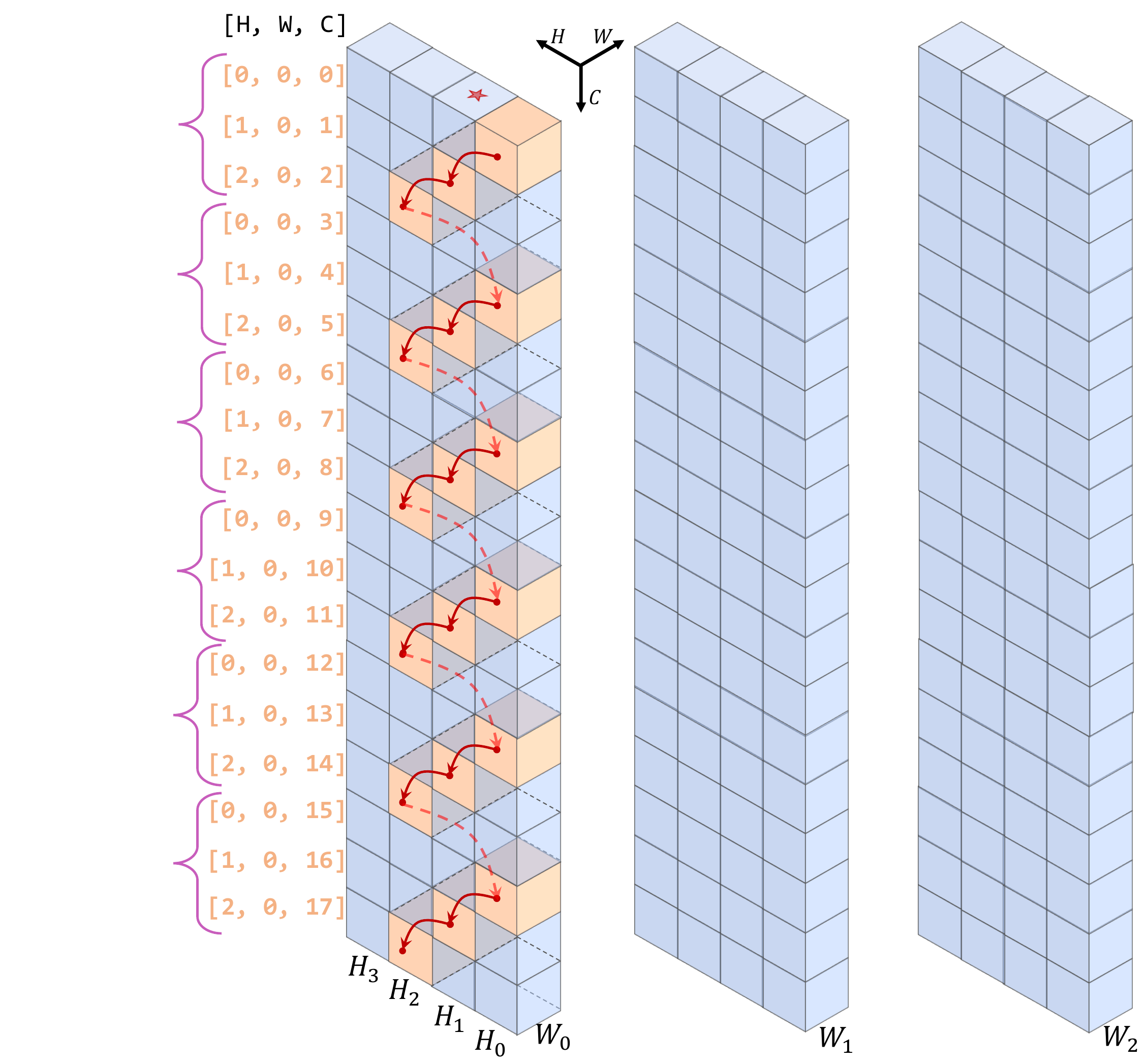}
    \vspace{-4mm}
    \caption{$S_H\times S_W$: $3\times1$}\label{fig:pattern3x1}
    \vspace{2mm}
\end{subfigure}
\hfill\vline\hfill
\begin{subfigure}{0.46\textwidth}
    \centering
    \includegraphics[width=\textwidth]{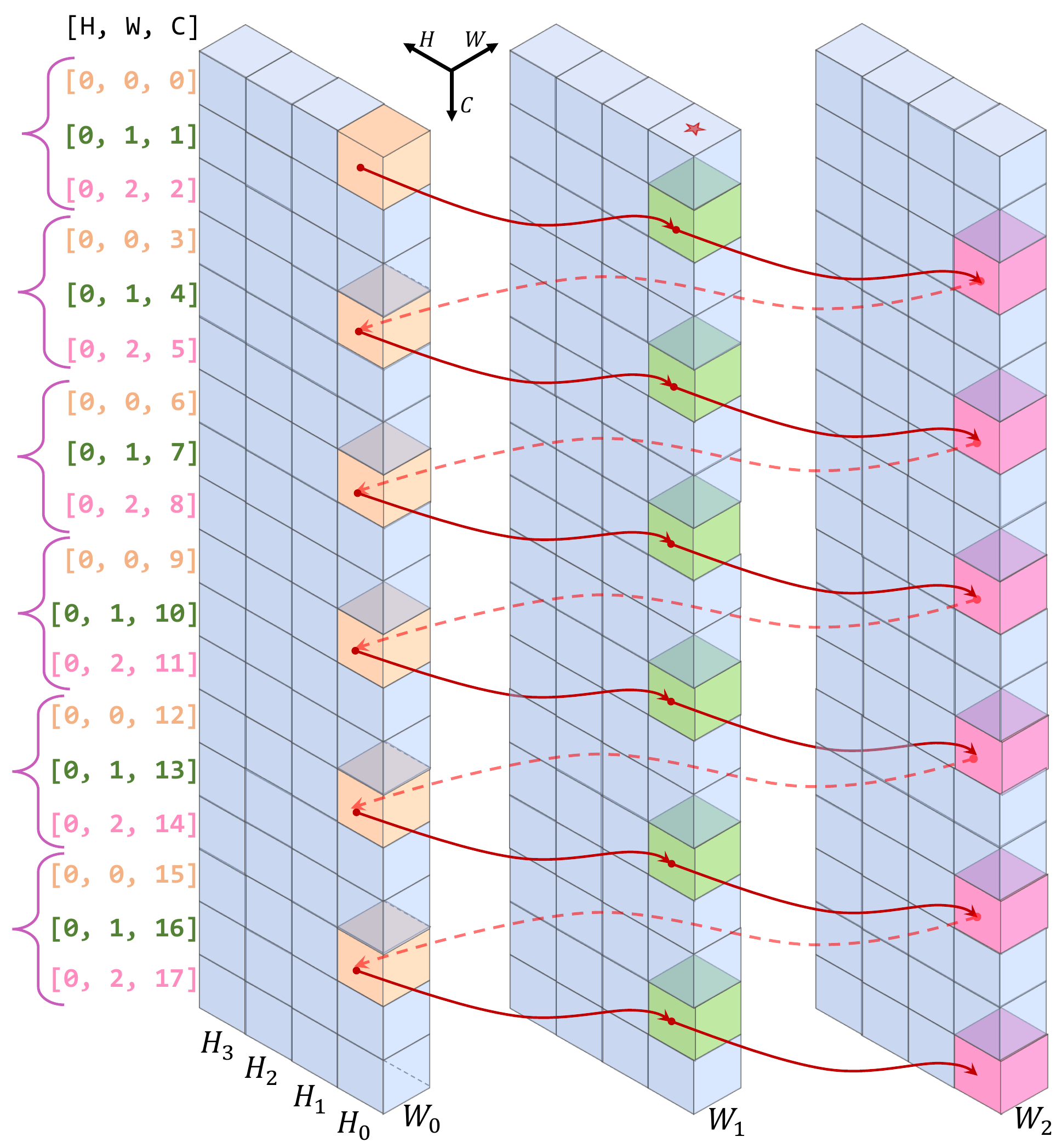}
    \vspace{-4mm}
    \caption{$S_H\times S_W$: $1\times3$}\label{fig:pattern1x3}
    \vspace{2mm}
\end{subfigure}
\caption{\rebuttal{\textbf{Examples of \texttt{Stepsize} cases:} Here we separate the feature map along the \texttt{width} dimension for convenient visualization.} \textcolor{redstar}{$\bigstar$} denotes the output position. We place the absolute coordinates~(\texttt{h, w, c}) of the sampled points at the left of the feature. Sampled points within a curly bracket~(\textbf{\textcolor{curly}{\{}}) belong to the same period (group). Dash lines link two cyclical periods. }\label{fig:general}
\end{figure}

\rebuttal{For easier understanding of our proposed CycleMLP, we visualize several instances of CycleMLP in Figure~\ref{fig:general}, including general case with stepsize 3$\times$3~(\ref{fig:pattern3x3}), even stepsize~(\ref{fig:pattern3x2}), and examples where stepsize along \texttt{height} or \texttt{width} equals to 1~(\ref{fig:pattern3x1}, \ref{fig:pattern1x3}).}

\rebuttal{We note that given specific number of input and output channels, no matter how the stepsize changes, the number of parameters of the CycleMLP does not change. Therefore, there is a trade-off of representation abilities between spatial and channel dimensions, which will be discussed in details in following experimental analysis.}

\rebuttal{\textbf{Experiments:} We further conduct experiments on CycleMLPs with stepsize of 2$\times$7, 7$\times$2, 7$\times$3, 3$\times$7, and 4$\times$4, respectively. The results are summarized Table~\ref{tab:even}. For fair comparisons, all the models in the above table have the same parameters and FLOPs. We observe that the model with stepsize of 1$\times$7 and 7$\times$1  achieves the best performance, especially for semantic segmentation on ADE20K. 
To analyze the impact of stepsize on the performance, we take Figure 7 for better illustration. One can see that enlarging the stepsize can expand the spatial receptive field. However, at a cost, it will reduce the number of periods (groups) running along the channel dimension, which may hurt the channel-wise representation abilities. Taking a feature map with $C=18$ for example, the CycleMLP with stepsize 3$\times$3 (Figure 7(a)) runs through only 2 channel groups (curly brackets in the figure). However, the CycleMLP with stepsize 3$\times$1 (Figure 7(c)) will run through 6 groups in total, making better use of the representation in the channel dimension. That’s to say, there is a trade-off between spatial and channel representation. We empirically found that CyCleMLP with stepsize of 1$\times$7 and 7$\times$1  achieves the best performance.
}

\end{document}